\documentclass[11pt,a4paper]{article}
\usepackage{times,latexsym}
\usepackage{url}
\usepackage[T1]{fontenc}
\usepackage{graphicx}
\usepackage{booktabs}
\usepackage{subfigure}
\usepackage{textcomp}
\usepackage{verbatim}
\usepackage{svg}
\usepackage{multirow}
\usepackage{array}
\usepackage{tabularx}
\usepackage{colortbl}
\usepackage{xcolor}
\usepackage{pdfpages}
\usepackage[margin=1in]{geometry}
\usepackage{xcolor}
\usepackage{array}
\usepackage{hyperref}
\usepackage{booktabs}
\usepackage{multirow}
\usepackage{enumitem}
\usepackage{CJKutf8}

\usepackage{appendix}
\usepackage{listings}
\usepackage{xcolor}   
\usepackage{tcolorbox} 
\definecolor{googlepurple}{RGB}{103, 58, 183}
\definecolor{lightyellow}{HTML}{FFFFCC}
\usepackage{multicol}
\definecolor{googledocsblue}{RGB}{66, 133, 244}

\definecolor{lightbrown}{RGB}{222, 184, 135}  
\definecolor{mediumbrown}{RGB}{210, 180, 140} 
\definecolor{darkbrown}{RGB}{139, 69, 19}     

\definecolor{hindi}{HTML}{FF69B4}       
\definecolor{indonesian}{HTML}{DAA520}  
\definecolor{japanese}{HTML}{228B22}    
\definecolor{vietnamese}{HTML}{20B2AA}  
\definecolor{tamil}{HTML}{1E90FF}       
\definecolor{PERcolor}{HTML}{FFF2CC}        
\definecolor{ORGcolor}{HTML}{CFE2F3}        
\definecolor{COUNTRYcolor}{HTML}{D9EAD3}    
\definecolor{LOCcolor}{HTML}{FCE5CD}        
\definecolor{EVENTcolor}{HTML}{EAD1DC}      
\definecolor{WORKcolor}{HTML}{D9D2E9}       
\definecolor{MISCcolor}{HTML}{EDEDED}       

\newcommand{\hlPER}[1]{\sethlcolor{PERcolor}\hl{#1}}
\newcommand{\hlORG}[1]{\sethlcolor{ORGcolor}\hl{#1}}
\newcommand{\hlCOUNTRY}[1]{\sethlcolor{COUNTRYcolor}\hl{#1}}
\newcommand{\hlLOC}[1]{\sethlcolor{LOCcolor}\hl{#1}}

\newcommand{\hlWORK}[1]{\sethlcolor{WORKcolor}\hl{#1}}
\newcommand{\hlMISC}[1]{\sethlcolor{MISCcolor}\hl{#1}}

   \usepackage[acceptedWithA]{tacl2021v1}
\usepackage{xspace,mfirstuc,tabulary}

\newif\iftaclinstructions
\taclinstructionsfalse 
\iftaclinstructions
\renewcommand{\confidential}{}
\renewcommand{\anonsubtext}{(No author info supplied here, for consistency with
TACL-submission anonymization requirements)}
\newcommand{\instr}
\fi

\iftaclpubformat 

\else

\fi

\title{MERLIN: A Testbed for \\ Multilingual Multimodal Entity Recognition and Linking}

\author{
  \textbf{Sathyanarayanan Ramamoorthy}$^\diamond$ \quad
  \textbf{Vishwa Shah}$^\diamond$ \quad
  \textbf{Simran Khanuja}$^\diamond$ \quad
  \textbf{Zaid Sheikh}$^\diamond$\\
  \textbf{Shan Jie}$^\dagger$\quad
  \textbf{Ann Chia}$^\dagger$ \quad
  \textbf{Shearman Chua}$^\dagger$ \quad
  \textbf{Graham Neubig}$^\diamond$ \\
  \
  $^\diamond$Carnegie Mellon University \\
  \texttt{\{sramamoo,vishwavs,skhanuja,zsheikh,gneubig\}@cs.cmu.edu} \\
  \
  $^\dagger$Defence Science and Technology Agency, Singapore \\
  \texttt{\{yshanjie,achiajin,schuawei\}@dsta.gov.sg}
}

\date{}

\usepackage{color-edits}
\usepackage{soul}
\usepackage{xcolor}
\sethlcolor{green!20}

\addauthor{gn}{magenta}

\begin{document}
\maketitle
\begin{abstract}

This paper introduces MERLIN, a novel testbed system for the task of \textit{Multilingual Multimodal Entity Linking}. The created dataset includes BBC news article titles, paired with corresponding images, in five languages: Hindi, Japanese, Indonesian, Vietnamese, and Tamil, featuring over 7,000 named entity mentions linked to 2,500 unique Wikidata entities. We also include several benchmarks using multilingual and multimodal entity linking methods exploring different language models like LLaMa-2 and Aya-23. Our findings indicate that incorporating visual data improves the accuracy of entity linking, especially for entities where the textual context is ambiguous or insufficient, and particularly for models that do not have strong multilingual abilities. For the work, the dataset, methods are available here \footnote{\url{https://github.com/rsathya4802/merlin}}.

\end{abstract}

\section{Introduction}

Entity Linking (EL) involves linking ambiguous entity mentions from unstructured data to their unambiguous referents in a knowledge base (KB). This is an important step in many downstream applications such as search \cite{Blanco2015FastAS}, question-answering \cite{fevry-etal-2020-entities, de2018question}, and sentiment analysis \cite{zhong2023knowledge}.
While traditional entity linking focused on text, in many settings (such as short news clips or social media posts), textual data alone can often be insufficient to adequately disambiguate entities given short, context-dependent, and ambiguous statements \citep{guo2013link}.
Images, which often accompany text, can provide crucial context for disambiguating entities \cite{Moon2018MultimodalNE, adjali2020building}.
Multiple prior works have proposed datasets \cite{wang-etal-2022-wikidiverse, gan2021multimodal} and methods \cite{zhang2021attention} for EL given visual and textual context.

Additionally, EL presents unique challenges on non-English languages \cite{fu2020design}, where resources are much more constrained.
In the multilingual context, earlier works framed this as a \emph{cross-lingual} problem, where entities in one language were linked to a KB in another language, typically English \cite{mcnamee-etal-2011-cross}. Subsequent works \cite{botha-etal-2020-entity, de2020autoregressive} utilized massively multilingual models with language-agnostic representations to link mentions to language-agnostic KBs in a multilingual setting. 

\begin{figure}[!ht]
    \centering
    \includegraphics[width=0.5\textwidth]{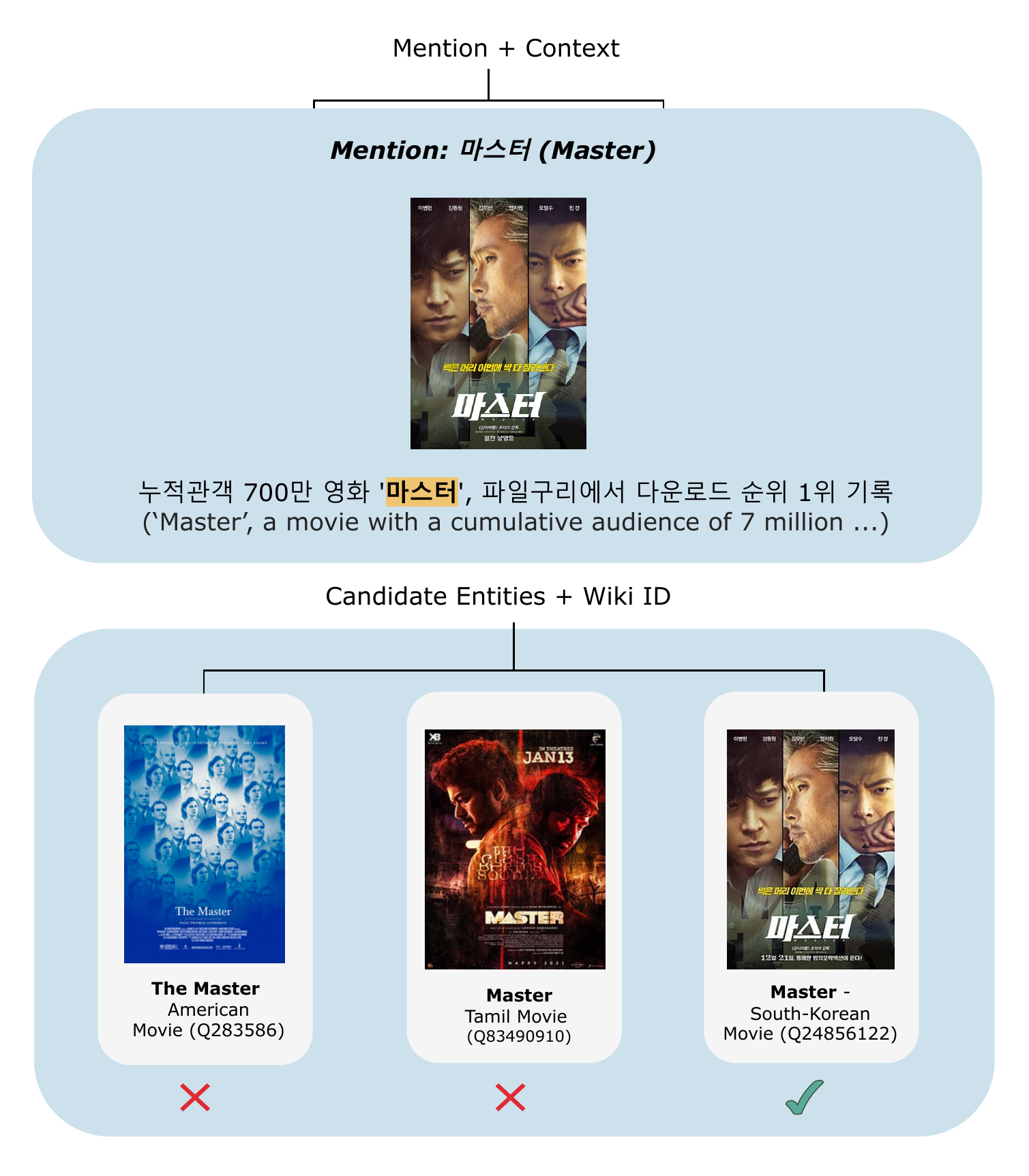}
    \caption{\textit{Resolving Entity Ambiguities Using Contextual and Visual Cues}. Entity disambiguation is challenging when multiple potential entities are associated with a single mention. Here, the "Master" films in different languages can each be linked to distinct IDs. Images serve as a vital tool allowing for correct linking of "Master" to its corresponding entity ID, thereby avoiding ambiguity.}
    
    \label{fig:images-help}
\end{figure}

Up until now, EL in the multilingual context has only dealt with the text modality. However, given the challenges of multilingual entity linking, it is natural to consider whether the visual modality may help resolve ambiguity in the multilingual case as well (as illustrated in Fig \ref{fig:images-help}). The novelty of our work revolves around formulating the task of multilingual entity linking through multimodal data along with a dataset and baseline methods, the goal being to show the performance of existing EL methods on our dataset and also to encourage the community to work on disambiguating multilingual entity mentions using images.
Therefore, to push the frontiers of EL towards this multilingual multimodal setting, we create \textbf{MERLIN}: the first testbed for \textbf{M}ultilingual Multimodal \textbf{E}ntity \textbf{R}ecognition and \textbf{LIN}king. MERLIN is a testset comprised of news article titles from BBC, paired with corresponding images. It features over 7,000 named entity mentions linked to 2,500 unique entities in the Wikidata knowledge base. The dataset covers five languages: Hindi, Japanese, Indonesian, Vietnamese, and Tamil.

In the following sections, we first formulate the task and describe the various components involved in building a \emph{multilingual multimodal entity linking} (\textbf{MMEL}) dataset in \S\ref{sec:task}. Next, in \S\ref{sec:dataset}, we describe the dataset selection and annotation process. In \S\ref{sec:baselines}, we evaluate several methods that have been proposed for multilingual or multimodal entity-linking \cite{mgenre, gemel}, on our newly created test set. Analysing results of these methods in \S\ref{sec:analysis}, we find that even the best approaches lack in performance, demonstrating the difficulty of our task and dataset. We also show that \emph{utilizing visual and text information} aids in disambiguating entities, leading to a significant improvement compared to methods that do not use visual information, especially for LLMs that aren't massively multilingual.
\section{Task Formulation}
\label{sec:task}

\textbf{Multilingual Multimodal Entity Linking (MMEL)} is the task of mapping mentions $m$ within multimodal and multilingual contexts to the corresponding entities in a knowledge base (KB). Formally, let $\mathcal{E}$ denote the entity set of the KB, typically comprising millions of entities $e$, each associated with a textual description $T_e$ and an image $V_e$. Each mention $m$ is characterized by its visual context $V_m$ and textual context $T_m$, where $V_m$ represents the image associated with $m$, and $T_m$ represents the textual spans surrounding $m$ and including $m$ (e.g., article title and associated image).\\ 

Given an image $V_m$, corresponding textual context $T_m$, and a mention $m$ embedded within $T_m$, the task of MMEL is to identify and link the mention $m$ to its corresponding entity $e \in \mathcal{E}$, where $\mathcal{E}$ is defined as the set of all Wikidata entities in the KB. Formally, the objective is to output a set of mention-entity pairs: $\{(m_i, e_i)\}_{i=1}^{n_m}$, where each entity $e_i$ corresponds to a mention $m_i$ and $n_m$ represents the total number of mentions in the given context.

In this multilingual setting, the context $T_m$ can be presented in various languages, thereby necessitating robust cross-lingual and cross-modal understanding. The assumption is that each mention $m_i$ has a valid corresponding entity $e_i$ within the KB, the so-called in-KB evaluation paradigm. We do not consider the challenge of out-of-KB predictions in this work.

\section{The making of MERLIN} \label{sec:dataset}

\subsection{Language and Dataset Selection}

We create a multimodal multilingual entity linking dataset in five languages, based on news articles from the M3LS dataset \cite{verma-etal-2023-large}, which was originally curated for the task of multilingual, multimodal summarization. The M3LS dataset comprises news articles with associated images covering a range of topics such as politics, sports, economy, science, and technology. These articles have been curated from the British Broadcasting Corporation (BBC) over a decade and span 20 languages, with 8 high-resource and 12 low-resource. 

The five languages selected for our study include Hindi, Japanese, Indonesian, Tamil, and Vietnamese. Our choice is guided by several factors: \textit{(i)} linguistic diversity, \textit{(ii)} speaker population, and \textit{(iii)} annotator availability.

\paragraph{Linguistic Diversity:} The selected languages represent a broad range of language families: Austronesian (\emph{Indonesian}), Austroasiatic (\emph{Vietnamese}), Indo-European (\emph{Hindi}), Dravidian (\emph{Tamil}), and Japonic (\emph{Japanese}). This diversity is further reflected in the various scripts used, including Devanagari (\emph{Hindi}), Kanji/Katakana/Hiragana (\emph{Japanese}), and Latin (\emph{Indonesian and Vietnamese}).

\paragraph{Speaker Population:} The chosen languages cover a wide range of the global population. Hindi is spoken by over 600 million people worldwide, Japanese by over 128 million, Indonesian by nearly 200 million (considering both native and second-language speakers), and Tamil and Vietnamese by approximately 75 million each. The inclusion of these languages in our dataset contributes to NLP research for a significant proportion of the non-English speaking global population \cite{blasi-etal-2022-systematic}.
Based on the number of articles in their dataset, M3LS classifies Tamil and Japanese as low-resource, but use Tamil and Indonesian in our dataset to advance NLP research in low-resource languages \cite{joshi-etal-2020-state}.

\paragraph{Annotator Availability:}
In order to create a high-quality dataset, it is also necessary that we can identify a sufficient population of annotators who can do high-quality annotation.
We utilized the Prolific Crowdsourcing Platform\footnote{\url{https://www.prolific.com/}} to recruit annotators, maintaining a threshold on the languages that have a minimum of 50 eligible participants. Eligibility was determined based on participants' nationality and native language, as required by our annotation design (Table \ref{tab:eligible-participants}).

To create a balanced dataset, we selected 1,000 article titles per language from M3LS, considering factors such as budget, annotator availability, and the time required for dataset curation. More details on the selection process will be released later. 

\begin{table}[t]
    \centering
    \small
    \begin{tabular}{ccc}
        \toprule
        \textbf{Nationality} & \textbf{\#Eligible} & \textbf{\#Selected} \\ \midrule
        India & 196 & 9\\ 
        Japan & 57 & 7\\ 
        Indonesia &  105 & 5\\ 
        India & 63 &6\\ 
        Vietnam & 105 & 6\\ 
        \bottomrule
    \end{tabular}
    \caption{Overview of eligible participants available in Prolific for the data annotation process, with a minimum threshold of 50 eligible participants}
    \label{tab:eligible-participants}
\end{table}

\subsection{Dataset Annotation} \label{sec:sec2.2}

For data annotation, we use the INCEpTION tool\footnote{\url{https://inception-project.github.io/}} \cite{klie-etal-2018-inception}, a web-based text-annotation platform designed for a range of tasks on written text, including Named Entity Recognition (NER), Part-of-Speech (POS) tagging, and Entity Linking. INCEpTION allows for the creation of projects through layers and tagsets, enabling customizable interfaces with support for both text and images. It also allows multiple collaborators to work simultaneously on the same annotation project and supports managing multiple projects concurrently. Notably, INCEpTION enables linking text to Knowledge Bases like Wikidata, which is essential for our work on entity linking, because we chose it as the knowledge base.

\paragraph{Prolific Recruitment and Pre-screening:} We used Prolific to recruit annotators based on their primary \textit{language} and \textit{nationality}, focusing on regions where the selected languages are most commonly spoken. To ensure the competence of the annotators, before they work on the annotation task, we implemented a two-stage process, first, a \textbf{pre-screening task} involving 10 articles in English with ground-truth labels from BBC articles, followed by the \textbf{main annotation task}. Annotators were required to achieve a minimum F1-score of 60\% in pre-screening to qualify for the main annotation task. This process ensured that only qualified annotators participated in the actual annotation, where we also relied on Prolific's filters to ensure their language proficiency.

\paragraph{Main Annotation Task:} The main annotation task involved 1,000 article titles per language, divided into 10 splits of 100 titles each to manage workload and maintain quality. Each split was annotated by \textbf{three different annotators}, and majority voting was used to determine the final entities for each mention. Annotators were provided with extensive guidelines and instructional videos to ensure consistent and accurate annotations. Further details on the number of annotators, guidelines, and links to videos used for providing instructions to annotators can be found in the appendix

\subsection{Dataset Analysis}

\begin{figure*}[t]
    \centering
    \includegraphics[width=1\textwidth]{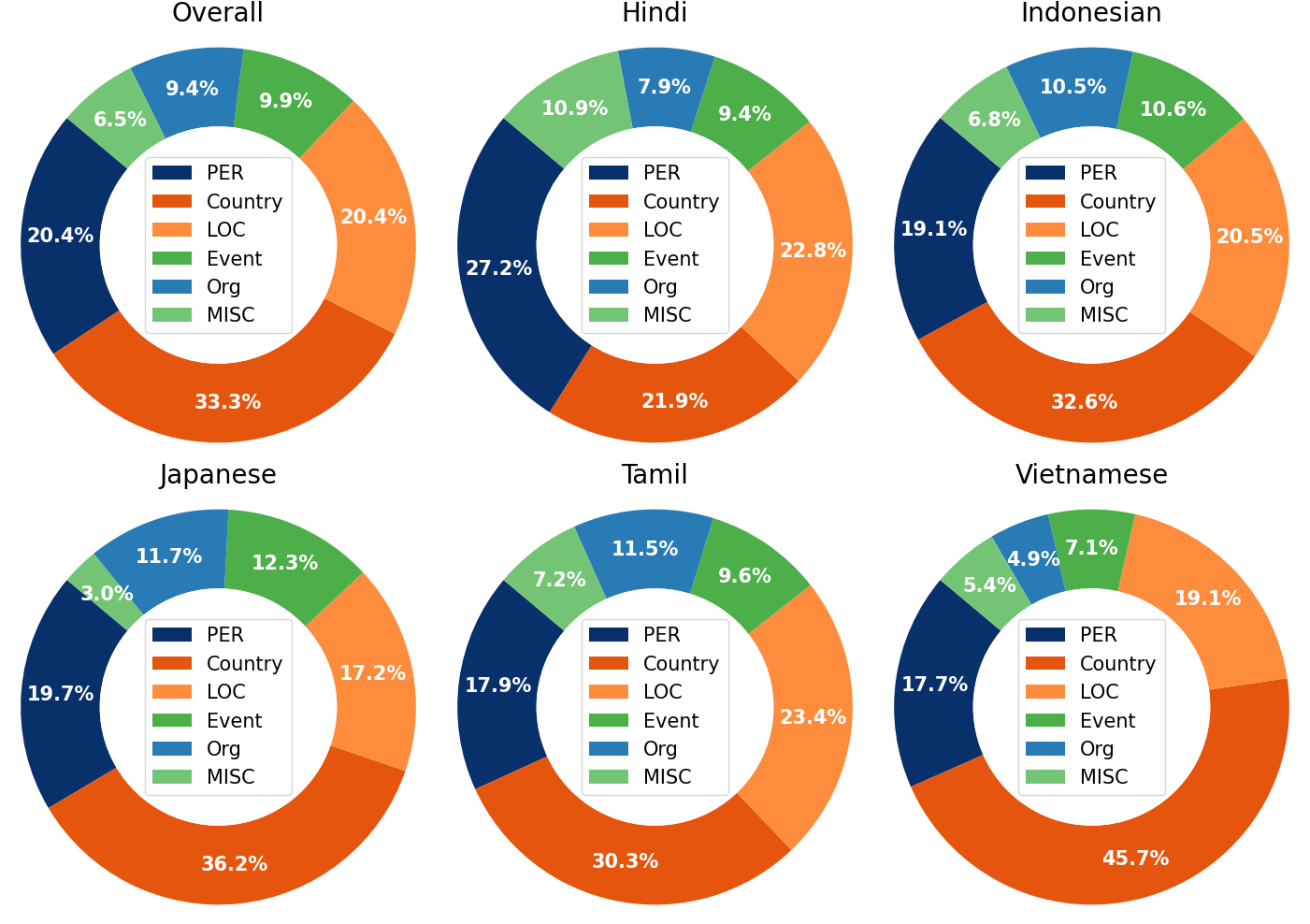}
    \caption{Category wise distribution of entities} 
    \label{fig:entity_category_plot}
\end{figure*}

\noindent \textbf{Entity Types} We analyze the distribution of entities based on the categories in Figure \ref{fig:entity_category_plot}. We use the "instance of" (P31) and "subclass of" (P279) properties to traverse the Wikidata hierarchy and identify the root entity, which is then mapped to either Person (PER), Country, Location (LOC), Organization (ORG), Event, or Miscellaneous (MISC). \\

\begin{table*}[ht!]
\centering
\begin{tabular}{lcccccc}
\toprule
 & Hindi & Indonesian & Japanese & Tamil & Vietnamese & Overall \\
\midrule
Number of article titles & 1000 & 1000 & 1000 & 1000 & 1000 & 5000 \\
Words per article title & 10.48 & 8.85 & 14.99 & 9.19 & 11.99 & 11.10 \\
Number of unique entities & 787 & 765 & 655 & 572 & 475 & 2480 \\
Number of mentions & 1480 & 1490 & 1720 & 1254 & 1343 & 7287 \\
Number of words per mention & 1.26 & 1.34 & 1.36 & 1.26 & 1.70 & 1.39 \\
Unlinked mentions* & 304 & 201 & 112 & 456 & 170 & 1243 \\
\bottomrule
\end{tabular}
\caption{Data and annotations per language, along with overall dataset statistics. Unlinked mentions (*) indicate the number of entity mentions that annotators could not link to a Wikidata page; these entities were excluded from methods.}
\label{tab:language_data}
\end{table*}

\noindent \textbf{Inter-annotator agreement}: Cohen's Kappa \cite{mchugh2012interrater} is a statistical measure used to evaluate the agreement between two annotators, taking into account the possibility of agreement occurring by chance. To calculate Cohen's Kappa for each language split between three annotators, we compare entity mentions annotated by each pair of annotators, two at a time. We first link the entity mentions marked by both annotators and focus on those where both made a selection. We then compare the Wikidata URLs chosen by the annotators to determine a match. This comparison forms the basis for calculating Cohen's Kappa, quantifying the level of agreement between annotators. The process is repeated for all annotator pairs across each language split to assess inter-annotator reliability. Table \ref{tab:kappa-results} shows average Cohen's Kappa values for inter-annotator agreement across five languages: Hindi, Indonesian, Japanese, Vietnamese, and Tamil. The values indicate the degree of consistency in entity linking between annotators, with most values suggesting almost perfect agreement.

\begin{table*}[ht]
\centering
\begin{tabular}{llclp{2.3cm}}
\toprule
\textbf{MEL Datasets} & \textbf{KB} & \textbf{Multilingual}  &\textbf{Multimodal} & \textbf{Languages} \\ \midrule
Mewsli \cite{botha-etal-2020-entity} & Wikidata& \colorbox{vietnamese}{Yes}&No& ar, de, en, es, fa, ja, sr, ta, tr\\ 
Twitter \cite{adjali2020building}              & Twitter Users & No                   &\colorbox{vietnamese}{Yes}& en                 \\ 
Zeshel \cite{logeswaran2019zero}& Wikia (Fandom)& No                     &No& en                 \\ 
WikiDiverse  \cite{wang-etal-2022-wikidiverse}         & Wikidata& No                     &\colorbox{vietnamese}{Yes}& en                 \\ \midrule
\textbf{MERLIN}       & \textbf{Wikidata} & \textbf{\colorbox{vietnamese}{Yes}}      &\textbf{\colorbox{vietnamese}{Yes}}& \textbf{hi, id, ja, ta, vi}\\ \bottomrule
\end{tabular}
\caption{Comparison between MERLIN and other EL datasets}
\label{tab:dataset-comp}
\end{table*}

\noindent \textbf{MERLIN v/s other EL datasets}: Unlike other datasets that mainly emphasize text-based entity linking and are predominantly monolingual (primarily English), MERLIN dataset combines both textual and visual data, drawing on Wikidata to cover a wide array of entity types ( Fig.  \ref{fig:entity_category_plot}) derived from news article titles. This multimodal, multilingual approach, enriched by manual curation, also incorporates the strengths of previous notable works in the entity linking domain, as illustrated in Table \ref{tab:dataset-comp}.

\begin{table}[ht]
\centering
\begin{tabular}{|c|c|}
\hline
\textbf{Language} & \textbf{Average Cohen's Kappa} \\
\hline
Hindi & 0.79 \\
Indonesian & 0.81 \\
Japanese & 0.91 \\
Vietnamese & 0.88 \\
Tamil & 0.76 \\
\hline
\textbf{Overall} & \textbf{0.83} \\
\hline
\end{tabular}
\caption{The agreement scores reflect the consistency among annotators, with Japanese and Vietnamese showing the highest agreement, while Tamil demonstrates greater variability. Scores indicate levels of agreement as follows: $\leq 0$ (no agreement), 0.01--0.20 (slight), 0.21--0.40 (fair), 0.41--0.60 (moderate), 0.61--0.80 (substantial), and 0.81--1.00 (almost perfect)}
\label{tab:kappa-results}
\end{table}

\section{Methods} \label{sec:baselines}

\begin{figure}[!ht]
    \centering
    \includegraphics[width=0.5\textwidth]{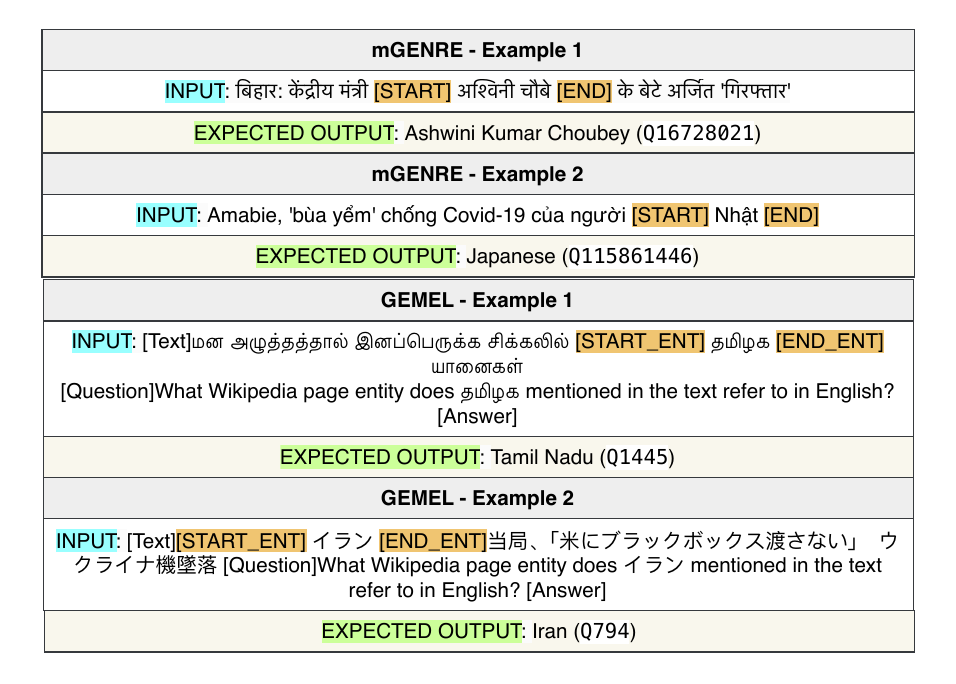}
    \caption{This figure illustrates examples of prompts used for mGENRE and GEMEL systems and how they are expected to resolve entity mentions in different languages by linking them to the correct entities}
    \label{fig:prompts}
\end{figure}

Traditional EL approaches typically employ a two-stage "retrieve and re-rank" strategy \cite{wu2019scalable, barba2022extend, lai2022improving, xu2022enhancing}. The first stage involves selecting potential entities from a large knowledge base. The second stage then re-ranks these candidates to disambiguate the final entity. However, this two-step approach can lead to compounded errors and reduced effectiveness. To address these limitations, generative approaches to entity-linking have recently been proposed, where entity names are directly generated auto-regressively. In this work, we consider two such methods: \textbf{i) mGENRE} \cite{mgenre} built in the multilingual context and \textbf{ii) GEMEL} \cite{gemel} built in the multimodal context. A brief overview of both methods is given below:

\textbf{mGENRE}: mGENRE uses a sequence-to-sequence architecture based on mBART with constrained decoding to autoregressively generate target entity names. It is a unimodal, text-only approach where it cross-encodes the text description $T_m$ and mention $m$ to generate the corresponding entity $e$ in the knowledge base $\mathcal{E}$. Constrained decoding enables efficient and fast search within a large knowledge base by restricting the search space to valid Wikipedia entity names, eliminating the need for extensive vector indices. The prompts we use are shown in Figure \ref{fig:prompts}. To summarize, mGENRE is a unimodal, text-only method, which takes the article title with the entity mention as input, and directly generates the identified KB entity name as output.

\textbf{GEMEL}: GEMEL is adapted to \textbf{MMEL} by mapping mentions $m$ within both visual $V_m$ and textual $T_m$ contexts to their corresponding entities $e$ in a knowledge base $\mathcal{E}$. The framework keeps the vision encoder and LLM frozen, and trains a feature mapper to enable cross-modal interactions along with in-context learning demonstrations for better generations. Their work primarily focuses on multimodal EL in English and uses Llama-2 (7B) as the text encoder. We similarly train GEMEL's visual feature mapper on Wikidiverse in English \cite{wang-etal-2022-wikidiverse} and test on MERLIN, zero-shot.

Given that Llama-2's \cite{Touvron2023Llama2O} pre-training data mostly consists of English tokens , we modify the architecture to include Aya-23 (8B) \cite{aryabumi2024aya23openweight} as the text encoder. Aya-23 is a multilingual LLM trained on diverse high-quality multilingual datasets, including Wikipedia. Llama-2 has limited multilingual capabilities, primarily focused on English, whereas Aya 23 is designed for strong performance across multiple languages, making it more effective in multilingual tasks.

\textbf{Knowledge Base for Entity Linking}: Both the baseline systems use Wikipedia titles as their knowledge bases. 
Constrained decoding for mGENRE is implemented through a prefix trie built on Wikipedia English titles comprising of almost 6M entities while we do not perform constrained decoding on the GEMEL methods. Since during annotation, our dataset is linked to a language-agnostic Wikidata KB, it is easy for us to preprocess the data to extract a mention's corresponding English Wikipedia titles and adapt them to these methods.

\begin{table*}[!t]
    \centering
\resizebox{\textwidth}{!}{
\begin{tabular}{lcccccccccc}
\toprule
\textbf{Pipeline} & \multicolumn{2}{c}{\textbf{Hindi 
 (\%)}} & \multicolumn{2}{c}{\textbf{Indonesian (\%)}} & \multicolumn{2}{c}{\textbf{Japanese (\%)}} & \multicolumn{2}{c}{\textbf{Vietnamese (\%)}} & \multicolumn{2}{c}{\textbf{Tamil (\%)}} \\
\cmidrule(lr){2-3} \cmidrule(lr){4-5} \cmidrule(lr){6-7} \cmidrule(lr){8-9} \cmidrule(lr){10-11}
& \textbf{T} & \textbf{T+V} & \textbf{T} & \textbf{T+V} & \textbf{T} & \textbf{T+V} & \textbf{T} & \textbf{T+V} & \textbf{T} & \textbf{T+V} \\
\midrule
\textbf{mGENRE} & 55.72 & - & \textbf{78.93} & - & 71.06 & - & \textbf{74.13} & - & \textbf{63.18} & - \\
\textbf{GEMEL-Llama-2} & 61.1 & 54.17 & 65.96 & 74.29 \textuparrow& 64.59 & \textbf{74.76} \textuparrow& 65.07 & 70.06 \textuparrow& 29.42 & 21.21 \\
\textbf{GEMEL-Aya-23} & 66.12 & \textbf{67.53} \textuparrow& 70.24 & 75.00 \textuparrow& 69.59 & 70.05 \textuparrow& 69.16 & 69.47 \textuparrow& 47.92 & 37.95 \\
\bottomrule
\end{tabular}
}
    \caption{Recall@1 Performance Across Different Language Pipelines. The table compares the  performance of various pipelines (mGENRE, GEMEL-Llama-2, GEMEL-Aya-23 8B, and their variants without images) across five languages: Hindi, Indonesian, Japanese, Vietnamese, and Tamil. (\textuparrow) refers to cases where visual modality leads to better performance}
    \label{tab:methods-results}
\end{table*}

\section{Results and Analysis} \label{sec:analysis}
The performance of various entity linking methods varies significantly across different languages and models, as summarized in Table~\ref{tab:methods-results}. To better understand these differences, we explore the following research questions.

\noindent \textbf{Q1) What impact does incorporating multimodal context have in low-resource languages?}

\noindent \textbf{Overall performance}: The mGENRE pipeline demonstrates strong performance, particularly in Indonesian, where it achieves the highest R@1 of 78.93\%. However, in other languages like Tamil, its performance is notably lower, with an R@1 of 63.18\%. Both Llama-2 and Aya are not trained on Tamil data and we notice consistent low performance in Tamil by the GEMEL methods. This can also be attributed to the scripts that Tamil and even Hindi use and the lack of ability of current methods to understand contexts in these languages.
 The GEMEL-Llama-2 7B model shows improvements over mGENRE in languages such as Japanese, where it achieves an R@1 of 74.76\%. The GEMEL - Aya-23 8B model also demonstrates competitive performance, particularly in Hindi and Tamil. In Hindi, it achieves an R@1 of 67.53\%, outperforming both mGENRE and GEMEL-Llama-2 as shown in Table \ref{tab:methods-results}. For languages like Hindi and Tamil, the performance of Aya-23 is better than Llama-2, though both are lesser than mGENRE, which was trained extensively on all languages, otherwise both models exhibit similar performance. \\

\begin{figure*}[t!]
    \centering
    \includegraphics[width=\linewidth]{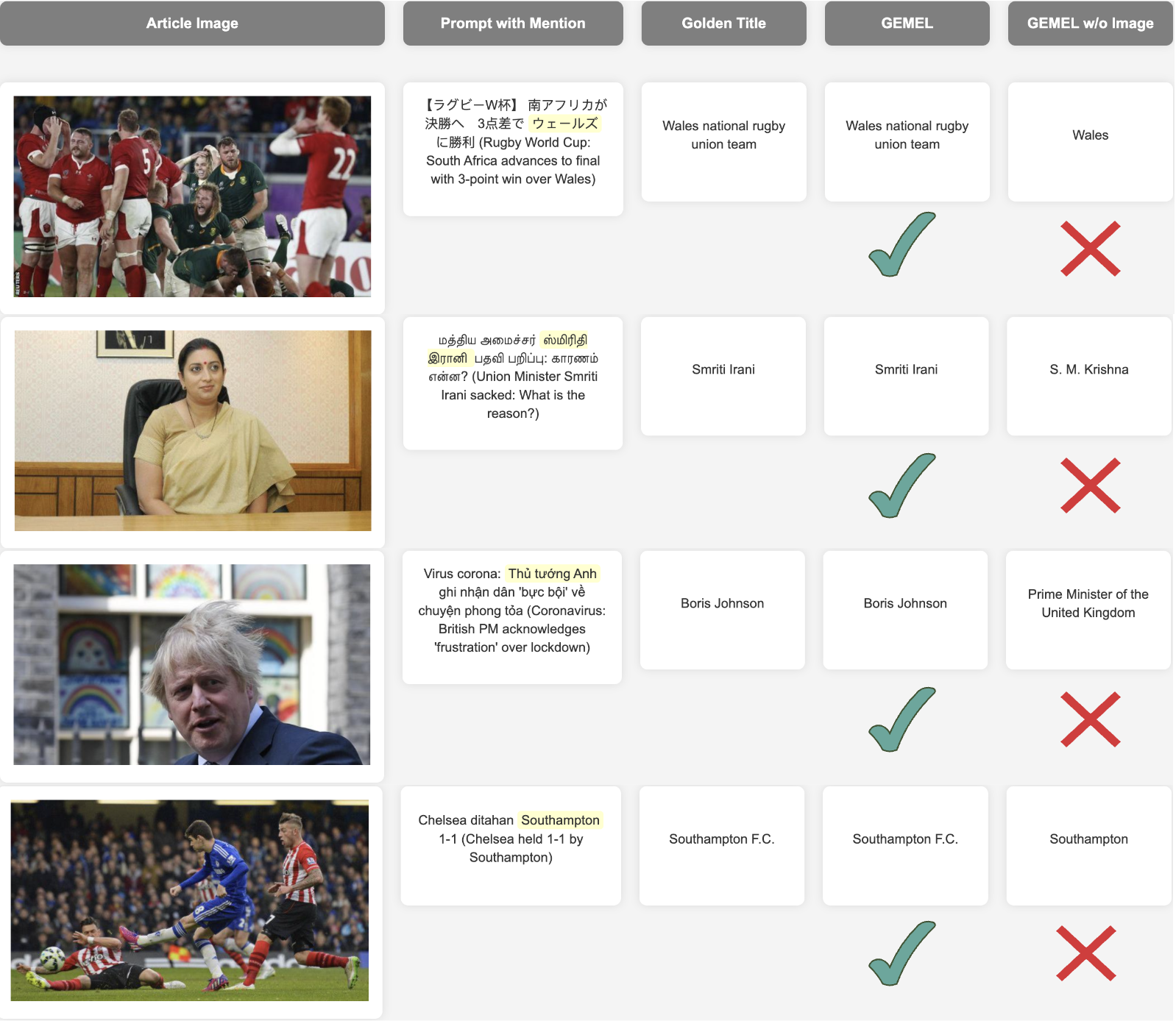}
    \caption{\textit{Impact of visual context in GEMEL}. The table compares the GEMEL model's predicted wikipedia titles with and without images across four examples. Each example includes the article image, prompt with \colorbox{lightyellow}{mention}, golden wikipedia title, and the predictions made by GEMEL with and without images}

    \label{fig:image_disambiguation}
\end{figure*}

\noindent \textbf{Images: The Secret Sauce?} We also compare our GEMEL's Llama-2 and Aya-23 variants, with and without images by introducing random noise instead of a visual prefix, to assess the significance of images in disambiguation. mGENRE and GEMEL are not directly comparable because both are fundamentally different architectures and mGENRE utilizes mBART as the text decoder, which is extensively multilingual and trained on over 100 languages. In contrast, Llama-2 is trained on significantly fewer multilingual datasets, and Aya-23's coverage includes only 23 languages, and lacks Tamil, a language present in MERLIN. Llama-2's performance \textbf{drops significantly} when images are removed, highlighting its strong reliance on visual data to achieve high accuracy. In contrast, Aya-23 experiences a much smaller drop without images, indicating that it is less dependent on visual input. This finding aligns with \citet{hu2023large}, who emphasized the importance of multimodality in cross-lingual tasks, particularly for models that lack explicit multilingual training. Overall, we hypothesize that image support is crucial for models that aren't trained to be multilingual, like Llama-2, but its importance reduces for models trained on diverse multilingual text, like Aya-23. While we cannot include visual context to mGENRE in the current setting, we believe that the multimodal gains in GEMEL will also apply to mGENRE, especially given it's strong multilingual capabilities.

A few qualitative examples of how images help in MMEL are shown in Fig \ref{fig:image_disambiguation}. In the first example, the visual context helps GEMEL correctly identify the \emph{Wales national rugby union team}, while the text-only model incorrectly links it to the country \emph{Wales}. In the next example, the visual context allows GEMEL to correctly identify \emph{Smriti Irani}, whereas the text-only model incorrectly links to \emph{S. M. Krishna}, another union minister in India. Similarly, in the third and fourth examples, the image aids in correctly identifying \emph{Southampton F.C.} and \emph{Boris Johnson}, while the text-only model fails by linking it to the city \emph{Southampton} and the position of \emph{Prime Minister of the United Kingdom}, respectively, which seem generic. \\

\begin{table*}[ht!]
    \centering
    \begin{tabular}{lccccc}
        \toprule
        & Hindi & Indonesian & Japanese & Vietnamese & Tamil \\
        \midrule
        Number of Examples & 112 & 54 & 50 & 203 & 77 \\
        \bottomrule
    \end{tabular}
    \caption{\textit{Number of ambiguous mentions across languages} idenitified by filtering mentions that can be linked to multiple Wikidata IDs.}

     \label{tab:ambiguous-count}
\end{table*}

\begin{table*}[ht!]
    \centering
    \begin{tabular}{lcccccc}
        \toprule
        Model & Modality & Hindi (\%) & Indonesian (\%)& Japanese (\%)& Vietnamese (\%)& Tamil (\%)\\
        \midrule
        \multirow{2}{*}{GEMEL-Llama-2} & T+V & 74.11 \textuparrow & 62.96 \textuparrow & 48.00 \textuparrow & 80.30 \textuparrow & 18.18 \\
        & T & 73.21 & 48.15 & 28.00 & 63.55 & 28.57 \\
        \midrule
        \multirow{2}{*}{GEMEL-Aya-23} & T+V & 75.00 & 66.67 & 48.00 \textuparrow & 67.00 \textuparrow & 38.96 \\
        & T & 75.00 & 66.67 & 46.00 & 59.61& 40.26 \\
        
        \bottomrule
    \end{tabular}
    \caption{\textbf{Accuracy scores across languages for ambiguous mentions}: For Hindi and Japanese, visual inputs provide marginal improvements, but they significantly help Indonesian and Vietnamese, (\textuparrow) indicates those. For Tamil, existing models struggle to get decent performance, even with visual inputs.}
    \label{tab:ambiguous-table}
\end{table*}

\noindent \textbf{Q2) What kind of instances especially benefit from visual inputs?}

\noindent \textbf{Ambiguous Mentions}:

Ambiguous mentions occur when the same textual mention can refer to multiple distinct Wikidata entities. These cases are particularly challenging as they require models to rely on context—sometimes beyond the text itself—to correctly associate the mention with the appropriate entity. We identify such mentions by filtering for instances where an annotated mention can be linked to multiple Wikidata IDs (e.g., 'Apple' can refer to both the fruit (\textit{Q89}) and the company (\textit{Q312}) in Wikidata). Qualitative examples from our dataset for such mentions can be seen in the Appendix Figure \ref{fig:ambig_entity_examples}. The total number of such instances identified are given in Table \ref{tab:ambiguous-count} and the accuracies of models across this subset are shown in Table \ref{tab:ambiguous-table}.

From these results, it is evident that performance is consistently lower on ambiguous mentions across all languages. With visual inputs, the gains are minimal but consistent, underscoring the difficulty of these ambiguous cases in the MERLIN dataset. Japanese and Tamil, in particular, show the most significant drop in accuracy, with Llama-2 scoring just 28\% and 18.18\% respectively in the text-only setting. This stark drop in performance suggests that current entity linking methods, even when enhanced with multimodal context, need further refinement to better handle ambiguity, especially in lower resourced languages.

\noindent \textbf{Influence of Mention Types}: We report performance across different mention types such as PER (Person), LOC (Location), Country, ORG (Organization), MISC (Miscellaneous), and Event in the appendix in Table \ref{tab:category-performance}. Overall, we observe that images significantly improve the accuracy for PER mentions across most languages, while ORG mentions are easier for models to resolve, with or without images. For MISC and Event mentions, we observe consistently lower accuracy scores across all models, suggesting that these types of mentions may require more contextual understanding that goes beyond the current multimodal approaches.

\noindent \textbf{Q3) How does performance of both models differ across languages?}

From Table \ref{tab:methods-results}, we observe that models like GEMEL-Llama-2 and GEMEL-Aya-23 generally perform better with the inclusion of visual context, across all languages. This suggests that images significantly aid in improving accuracy, particularly in low-resource settings.
We also notice that both Llama-2 and Aya-23 struggle on both Hindi and Tamil, while even though mGENRE is trained on these languages it's performance is relatively low. Except for GEMEL-Llama vs mGENRE for Japanese and GEMEL-Llama-unimodal vs mGENRE for Hindi, there are statistically significant differences between both systems (p $<$ 0.05). Overall, while visual context consistently boosts performance across languages, its impact is more critical in certain languages and categories, underlining the importance of multimodal approaches in enhancing entity linking accuracy, while also showing the amount of future research that needs to be explored in such languages.

\noindent \textbf{Q4) To what extent does translating to English help improve performance on MERLIN?}

To understand the impact of language on MERLIN's performance, we translated all non-English data into English using the NLLB model \cite{costa2022no}. By comparing the English and multilingual results, we aimed to identify any gaps in understanding. This process allowed us to assess the capabilities of GEMEL-Llama-2-7B and Aya-23 8B and identify any potential biases towards the English language.
The translation process led to a reduction in the number of entities due to translation inaccuracies. As a result, the evaluation was performed in a reduced entity space, focusing on those entities that were reliably translated. 
When translating multilingual data to English, we filtered out entities whose mentions did not appear in both the source and translated text. This filtering was necessary because translation can alter named entities through omission, transliteration differences, cultural adaptations, or structural sentence changes. By evaluating on this filtered subset, we ensured that performance differences between versions reflected the models' language capabilities rather than translation artifacts. We evaluated the models on the translated datasets to understand their performance in this limited space. A comparison of the performance of GEMEL-Llama-2-7B and Aya-23 8B across the five languages can be seen in Table \ref{tab:performance_comparison}.
We observe a general improvement in performance across all languages, underscoring the disparity in understanding between English and non-English contexts in these methods. Notably, Tamil and Hindi show significant gains.

\begin{table}[h!]
    \centering
    
    \resizebox{0.5\textwidth}{!}{
    \begin{tabular}{cccccc}
        \toprule
         \textbf{Model} & \textbf{Hindi} & \textbf{Indonesian} & \textbf{Japanese} & \textbf{Vietnamese} & \textbf{Tamil} \\
        \midrule
         Llama-2 & 77.57& 85.43& 81.72& 84.53& 80.19\\
         Aya-23 & 76.68 & 82.83 & 79.69 & 82.13 & 77.87 \\
         Llama-2* & 62.55& 80.75& 86.29& 78.82& 33.61\\
         Aya-23* & 71.86& 80.88& 84.77& 74.62& 42.76\\
        
        \bottomrule
    \end{tabular}
    }
    \caption{R@1 (in \%) Performance Comparison on Translated Data for GEMEL methods. * represents performance of models on original untranslated dataset}
    \label{tab:performance_comparison}
\end{table}

\section{Related Work}
The task of Entity Linking (EL) has evolved significantly, adapting to new challenges across various domains, including multilinguality, multimodality, and the utilization of diverse knowledge bases (KBs).

\noindent \textbf{Multilinguality in Entity Linking}: Entity Linking in a multilingual context has seen substantial advancements, particularly with the introduction of models that can handle multiple languages and link entities across different language KBs. Early work in this area often focused on cross-lingual EL, where mentions in one language were linked to entities in another, typically English, KBs. For example, \citet{mcnamee-etal-2011-cross} and Tsai and Roth (2016b) addressed the cross-lingual EL problem, while \cite{sil-florian-2016-one} and \citet{zhou2020improving} explored zero-shot transfer and low-resource languages.

With the advent of large language models (LLMs), approaches shifted towards more scalable and generalizable methods. \citet{botha-etal-2020-entity} introduced a multilingual EL model that extended bi-encoder architectures to over 100 languages, enabling linking to a language-agnostic reference KB. This was a significant departure from earlier methods that were limited to English KBs. The mGENRE \cite{mgenre} model  further advanced the field by using an autoregressive approach, which linked entities by generating unique tags in a multilingual setting. This method allowed for linking entities across languages without relying on monolingual KBs, marking a significant step forward in multilingual EL.

Recent works, such as BELA \cite{plekhanov2023multilingual}, expanded the scope by incorporating dense retrieval models that support a wide range of languages, including those with limited resources. These models leverage advanced neural architectures to address challenges such as domain adaptation and zero-shot learning, which are critical for multilingual EL in diverse and dynamic environments. \\

\noindent \textbf{Multimodality in EL:} The emergence of multimodal content, particularly in social media and multimedia platforms, has driven the development of multimodal EL (MEL). \cite{Moon2018MultimodalNE} pioneered the MEL task, introducing a zero-shot framework that combined textual, visual, and lexical information to link entities in social media posts. However, challenges such as the availability of datasets due to privacy regulations (e.g., GDPR) and the limited scope of existing datasets have impeded progress in this area.

Subsequent efforts have focused on creating more comprehensive and challenging MEL datasets. For instance, \cite{adjali2020building} developed a Twitter-based MEL dataset, although it was limited by entity types and ambiguity. Similarly, \cite{zhang2021attention} compiled a Chinese MEL dataset focusing on person entities, while \cite{gan2021multimodal} collected MEL data from movie reviews, targeting characters and persons in the movie domain.

Recent advancements in MEL include the DWE+ model \cite{song2024dwe+}, which addresses semantic inconsistencies by employing hierarchical contrastive learning and fine-grained feature extraction. AMELI \cite{yao2023ameli}introduces attribute-aware MEL, integrating structured attributes into the disambiguation process, while DRIN \cite{Xing2023DRINDR} enhances alignment through a dynamic Graph Convolutional Network, significantly improving fine-grained alignment across multimodal data.

\noindent\textbf{Multilingual, Multimodal Datasets}: The intersection of multilingual and multimodal EL represents a new frontier in EL research. Recent work such as 2M-NER \cite{wang20242m} demonstrates the potential of incorporating multimodal data into multilingual Named Entity Recognition (NER) tasks, paving the way for more comprehensive multilingual multimodal EL systems. This approach aligns closely with the objectives of our dataset, which seeks to bridge the gap between multilinguality and multimodality by providing a rich, annotated dataset that spans multiple languages and modalities.

Our contribution to this growing field is a multilingual multimodal EL dataset that supports the linking of entities across diverse languages and modalities, offering a robust benchmark for evaluating future EL models.

\section{Conclusion}

In this paper, we introduced MERLIN, a novel multilingual, multimodal entity linking dataset encompassing five diverse languages and scripts. The dataset, curated from BBC news articles with associated images, presents a unique opportunity to study the impact of visual information on entity linking across various languages. We explored several baseline systems, using language models like mGENRE, LLama-2 and Aya-23B, to establish a benchmark for this challenging task. Our analysis highlights the significant benefits of utilizing visual cues, particularly for disambiguating entities with ambiguous textual descriptions. However, the results also underscore the inherent difficulty of entity linking, especially in a multilingual context on low-resourced languages.
\section{Limitations}

Though MERLIN offers a valuable resource for multilingual multimodal entity linking research, it does have some limitations as follows,\\ 

\noindent \textbf{Different writing styles and limited genre}: The current data set is drawn exclusively from BBC news articles. We tried to make it as diverse as possible, considering news types, categories, etc. but including data from domains beyond news articles would have ensured a more comprehensive evaluation of model generalization abilities to different writing styles and entity distributions. This limited genre diversity may affect the generalizability of models trained on our dataset to broader real-world applications and potentially constrains the dataset's ability to benchmark state-of-the-art approaches in more diverse scenarios.\\ 

\noindent \textbf{Annotator Skill and Knowledge}: Although our Cohen's Kappa values indicate strong inter-annotator agreement, the data may still contain inconsistencies due to variations in annotators' language proficiencies, general knowledge, annotation skills and inevitable human errors.\\ 

\noindent \textbf{Wikipedia based baseline methods}: Finally, while our chosen baseline methods show good results, they link to Wikipedia page titles rather than directly to Wikidata QIDs, even though our dataset supports Wikidata linking. So they may not directly reflect the performance on a pure Wikidata QID linking task, which needs to be explored through better modeling strategies.

\section{Acknowledgments}

We would like to thank Yash Butala and Raghav Kapoor for their valuable contributions to the foundational work of MERLIN. This work was supported by a research grant from Defense Science and Technology Agency (DSTA), Singapore.

\bibliography{tacl2021}

\begin{thebibliography}{37}
\expandafter\ifx\csname natexlab\endcsname\relax\def\natexlab#1{#1}\fi

\bibitem[{Adjali et~al.(2020)Adjali, Besan{\c{c}}on, Ferret, Le~Borgne, and Grau}]{adjali2020building}
Omar Adjali, Romaric Besan{\c{c}}on, Olivier Ferret, Herv{\'e} Le~Borgne, and Brigitte Grau. 2020.
\newblock \href {https://aclanthology.org/2020.lrec-1.528} {Building a multimodal entity linking dataset from tweets}.
\newblock In \emph{Proceedings of the Twelfth Language Resources and Evaluation Conference}, pages 4285--4292, Marseille, France. European Language Resources Association.

\bibitem[{Aryabumi et~al.(2024)Aryabumi, Dang, Talupuru, Dash, Cairuz, Lin, Venkitesh, Smith, Campos, Tan, Marchisio, Bartolo, Ruder, Locatelli, Kreutzer, Frosst, Gomez, Blunsom, Fadaee, Üstün, and Hooker}]{aryabumi2024aya23openweight}
Viraat Aryabumi, John Dang, Dwarak Talupuru, Saurabh Dash, David Cairuz, Hangyu Lin, Bharat Venkitesh, Madeline Smith, Jon~Ander Campos, Yi~Chern Tan, Kelly Marchisio, Max Bartolo, Sebastian Ruder, Acyr Locatelli, Julia Kreutzer, Nick Frosst, Aidan Gomez, Phil Blunsom, Marzieh Fadaee, Ahmet Üstün, and Sara Hooker. 2024.
\newblock \href {http://arxiv.org/abs/2405.15032v2} {Aya 23: Open weight releases to further multilingual progress}.
\newblock \emph{arXiv preprint arXiv:2405.15032v2}.

\bibitem[{Barba et~al.(2022)Barba, Procopio, and Navigli}]{barba2022extend}
Edoardo Barba, Luigi Procopio, and Roberto Navigli. 2022.
\newblock \href {https://doi.org/10.18653/v1/2022.acl-long.177} {{E}xt{E}n{D}: Extractive entity disambiguation}.
\newblock In \emph{Proceedings of the 60th Annual Meeting of the Association for Computational Linguistics (Volume 1: Long Papers)}, pages 2478--2488, Dublin, Ireland. Association for Computational Linguistics.

\bibitem[{Blanco et~al.(2015)Blanco, Ottaviano, and Meij}]{Blanco2015FastAS}
Roi Blanco, Giuseppe Ottaviano, and Edgar Meij. 2015.
\newblock \href {https://doi.org/10.1145/2684822.2685317} {Fast and space-efficient entity linking for queries}.
\newblock In \emph{Proceedings of the Eighth ACM International Conference on Web Search and Data Mining}, WSDM '15, page 179–188, New York, NY, USA. Association for Computing Machinery.

\bibitem[{Blasi et~al.(2022)Blasi, Anastasopoulos, and Neubig}]{blasi-etal-2022-systematic}
Damian Blasi, Antonios Anastasopoulos, and Graham Neubig. 2022.
\newblock \href {https://doi.org/10.18653/v1/2022.acl-long.376} {Systematic inequalities in language technology performance across the world{'}s languages}.
\newblock In \emph{Proceedings of the 60th Annual Meeting of the Association for Computational Linguistics (Volume 1: Long Papers)}, pages 5486--5505, Dublin, Ireland. Association for Computational Linguistics.

\bibitem[{Botha et~al.(2020)Botha, Shan, and Gillick}]{botha-etal-2020-entity}
Jan~A. Botha, Zifei Shan, and Daniel Gillick. 2020.
\newblock \href {https://doi.org/10.18653/v1/2020.emnlp-main.630} {{E}ntity {L}inking in 100 {L}anguages}.
\newblock In \emph{Proceedings of the 2020 Conference on Empirical Methods in Natural Language Processing (EMNLP)}, pages 7833--7845, Online. Association for Computational Linguistics.

\bibitem[{Costa-juss{\`a} et~al.(2022)Costa-juss{\`a}, Cross, {\c{C}}elebi, Elbayad, Heafield, Heffernan, Kalbassi, Lam, Licht, Maillard et~al.}]{costa2022no}
Marta~R Costa-juss{\`a}, James Cross, Onur {\c{C}}elebi, Maha Elbayad, Kenneth Heafield, Kevin Heffernan, Elahe Kalbassi, Janice Lam, Daniel Licht, Jean Maillard, et~al. 2022.
\newblock No language left behind: Scaling human-centered machine translation.
\newblock \emph{arXiv preprint arXiv:2207.04672}.

\bibitem[{De~Cao et~al.(2019)De~Cao, Aziz, and Titov}]{de2018question}
Nicola De~Cao, Wilker Aziz, and Ivan Titov. 2019.
\newblock \href {https://doi.org/10.18653/v1/N19-1240} {Question answering by reasoning across documents with graph convolutional networks}.
\newblock In \emph{Proceedings of the 2019 Conference of the North {A}merican Chapter of the Association for Computational Linguistics: Human Language Technologies, Volume 1 (Long and Short Papers)}, pages 2306--2317, Minneapolis, Minnesota. Association for Computational Linguistics.

\bibitem[{De~Cao et~al.(2021)De~Cao, Izacard, Riedel, and Petroni}]{de2020autoregressive}
Nicola De~Cao, Gautier Izacard, Sebastian Riedel, and Fabio Petroni. 2021.
\newblock \href {https://arxiv.org/abs/2010.00904v3} {Autoregressive entity retrieval}.
\newblock \emph{arXiv preprint arXiv:2010.00904v3}.

\bibitem[{De~Cao et~al.(2022)De~Cao, Wu, Popat, Artetxe, Goyal, Plekhanov, Zettlemoyer, Cancedda, Riedel, and Petroni}]{mgenre}
Nicola De~Cao, Ledell Wu, Kashyap Popat, Mikel Artetxe, Naman Goyal, Mikhail Plekhanov, Luke Zettlemoyer, Nicola Cancedda, Sebastian Riedel, and Fabio Petroni. 2022.
\newblock \href {https://doi.org/10.1162/tacl_a_00460} {Multilingual autoregressive entity linking}.
\newblock \emph{Transactions of the Association for Computational Linguistics}, 10:274--290.

\bibitem[{F{\'e}vry et~al.(2020)F{\'e}vry, Baldini~Soares, FitzGerald, Choi, and Kwiatkowski}]{fevry-etal-2020-entities}
Thibault F{\'e}vry, Livio Baldini~Soares, Nicholas FitzGerald, Eunsol Choi, and Tom Kwiatkowski. 2020.
\newblock \href {https://doi.org/10.18653/v1/2020.emnlp-main.400} {Entities as experts: Sparse memory access with entity supervision}.
\newblock In \emph{Proceedings of the 2020 Conference on Empirical Methods in Natural Language Processing (EMNLP)}, pages 4937--4951, Online. Association for Computational Linguistics.

\bibitem[{Fu et~al.(2020)Fu, Shi, Yu, Zhao, and Roth}]{fu2020design}
Xingyu Fu, Weijia Shi, Xiaodong Yu, Zian Zhao, and Dan Roth. 2020.
\newblock Design challenges in low-resource cross-lingual entity linking.
\newblock \emph{arXiv preprint arXiv:2005.00692}.

\bibitem[{Gan et~al.(2021)Gan, Luo, Wang, Wang, He, and Huang}]{gan2021multimodal}
Jingru Gan, Jinchang Luo, Haiwei Wang, Shuhui Wang, Wei He, and Qingming Huang. 2021.
\newblock \href {https://doi.org/10.1145/3474085.3475400} {Multimodal entity linking: A new dataset and a baseline}.
\newblock In \emph{Proceedings of the 29th ACM International Conference on Multimedia}, MM '21, page 993–1001, New York, NY, USA. Association for Computing Machinery.

\bibitem[{Guo et~al.(2013)Guo, Chang, and Kiciman}]{guo2013link}
Stephen Guo, Ming-Wei Chang, and Emre Kiciman. 2013.
\newblock \href {https://aclanthology.org/N13-1122} {To link or not to link? a study on end-to-end tweet entity linking}.
\newblock In \emph{Proceedings of the 2013 Conference of the North {A}merican Chapter of the Association for Computational Linguistics: Human Language Technologies}, pages 1020--1030, Atlanta, Georgia. Association for Computational Linguistics.

\bibitem[{Hu et~al.(2023)Hu, Yao, Wang, Wang, Pan, Chen, Yu, Wu, Zhao, Zhang, Han, Lin, Xue, Li, Liu, and Sun}]{hu2023large}
Jinyi Hu, Yuan Yao, Chongyi Wang, Shan Wang, Yinxu Pan, Qianyu Chen, Tianyu Yu, Hanghao Wu, Yue Zhao, Haoye Zhang, Xu~Han, Yankai Lin, Jiao Xue, Dahai Li, Zhiyuan Liu, and Maosong Sun. 2023.
\newblock \href {https://arxiv.org/abs/2308.12038v3} {Large multilingual models pivot zero-shot multimodal learning across languages}.
\newblock \emph{arXiv preprint arXiv:2308.12038v3}.

\bibitem[{Joshi et~al.(2020)Joshi, Santy, Budhiraja, Bali, and Choudhury}]{joshi-etal-2020-state}
Pratik Joshi, Sebastin Santy, Amar Budhiraja, Kalika Bali, and Monojit Choudhury. 2020.
\newblock \href {https://doi.org/10.18653/v1/2020.acl-main.560} {The state and fate of linguistic diversity and inclusion in the {NLP} world}.
\newblock In \emph{Proceedings of the 58th Annual Meeting of the Association for Computational Linguistics}.

\bibitem[{Klie et~al.(2018)Klie, Bugert, Boullosa, Eckart~de Castilho, and Gurevych}]{klie-etal-2018-inception}
Jan-Christoph Klie, Michael Bugert, Beto Boullosa, Richard Eckart~de Castilho, and Iryna Gurevych. 2018.
\newblock \href {https://www.aclweb.org/anthology/C18-2002} {The {INCE}p{TION} platform: Machine-assisted and knowledge-oriented interactive annotation}.
\newblock In \emph{Proceedings of the 27th International Conference on Computational Linguistics: System Demonstrations}, pages 5--9, Santa Fe, New Mexico.

\bibitem[{Lai et~al.(2022)Lai, Ji, and Zhai}]{lai2022improving}
Tuan Lai, Heng Ji, and ChengXiang Zhai. 2022.
\newblock \href {https://doi.org/10.18653/v1/2022.findings-acl.292} {Improving candidate retrieval with entity profile generation for {W}ikidata entity linking}.
\newblock In \emph{Findings of the Association for Computational Linguistics: ACL 2022}, pages 3696--3711, Dublin, Ireland. Association for Computational Linguistics.

\bibitem[{Logeswaran et~al.(2019)Logeswaran, Chang, Lee, Toutanova, Devlin, and Lee}]{logeswaran2019zero}
Lajanugen Logeswaran, Ming-Wei Chang, Kenton Lee, Kristina Toutanova, Jacob Devlin, and Honglak Lee. 2019.
\newblock \href {https://doi.org/10.18653/v1/P19-1335} {Zero-shot entity linking by reading entity descriptions}.
\newblock In \emph{Proceedings of the 57th Annual Meeting of the Association for Computational Linguistics}, pages 3449--3460, Florence, Italy. Association for Computational Linguistics.

\bibitem[{McHugh(2012)}]{mchugh2012interrater}
Mary~L McHugh. 2012.
\newblock Interrater reliability: the kappa statistic.
\newblock \emph{Biochemia medica}, 22(3):276--282.

\bibitem[{McNamee et~al.(2011)McNamee, Mayfield, Lawrie, Oard, and Doermann}]{mcnamee-etal-2011-cross}
Paul McNamee, James Mayfield, Dawn Lawrie, Douglas Oard, and David Doermann. 2011.
\newblock \href {https://aclanthology.org/I11-1029} {Cross-language entity linking}.
\newblock In \emph{Proceedings of 5th International Joint Conference on Natural Language Processing}, pages 255--263, Chiang Mai, Thailand. Asian Federation of Natural Language Processing.

\bibitem[{Moon et~al.(2018)Moon, Neves, and Carvalho}]{Moon2018MultimodalNE}
Seungwhan Moon, Leonardo Neves, and Vitor~R. Carvalho. 2018.
\newblock \href {https://api.semanticscholar.org/CorpusID:51882074} {Multimodal named entity disambiguation for noisy social media posts}.
\newblock In \emph{Annual Meeting of the Association for Computational Linguistics}.

\bibitem[{Plekhanov et~al.(2023)Plekhanov, Kassner, Popat, Martin, Merello, Kozlovskii, Dreyer, and Cancedda}]{plekhanov2023multilingual}
Mikhail Plekhanov, Nora Kassner, Kashyap Popat, Louis Martin, Simone Merello, Borislav Kozlovskii, Frédéric~A. Dreyer, and Nicola Cancedda. 2023.
\newblock \href {http://arxiv.org/abs/2306.08896v1} {Multilingual end to end entity linking}.
\newblock \emph{arXiv preprint arXiv:2306.08896v1}.

\bibitem[{Shi et~al.(2024)Shi, Xu, Hu, and Zhang}]{gemel}
Senbao Shi, Zhenran Xu, Baotian Hu, and Min Zhang. 2024.
\newblock \href {https://aclanthology.org/2024.lrec-main.676} {Generative multimodal entity linking}.
\newblock In \emph{Proceedings of the 2024 Joint International Conference on Computational Linguistics, Language Resources and Evaluation (LREC-COLING 2024)}, pages 7654--7665, Torino, Italia. ELRA and ICCL.

\bibitem[{Sil and Florian(2016)}]{sil-florian-2016-one}
Avirup Sil and Radu Florian. 2016.
\newblock \href {https://doi.org/10.18653/v1/P16-1213} {One for all: Towards language independent named entity linking}.
\newblock In \emph{Proceedings of the 54th Annual Meeting of the Association for Computational Linguistics (Volume 1: Long Papers)}, pages 2255--2264, Berlin, Germany. Association for Computational Linguistics.

\bibitem[{Song et~al.(2024)Song, Li, Zhao, Li, Wang, Yu, Ma, Yan, Ji, and Mao}]{song2024dwe+}
Shezheng Song, Shasha Li, Shan Zhao, Xiaopeng Li, Chengyu Wang, Jie Yu, Jun Ma, Tianwei Yan, Bin Ji, and Xiaoguang Mao. 2024.
\newblock \href {https://arxiv.org/abs/2404.04818v1} {{DWE+}: Dual-way matching enhanced framework for multimodal entity linking}.
\newblock \emph{arXiv preprint arXiv:2404.04818v1}.

\bibitem[{Touvron et~al.(2023)Touvron, Martin, Stone, Albert, Almahairi, Babaei, Bashlykov, Batra, Bhargava, Bhosale, Bikel, Blecher, Ferrer, Chen, Cucurull, Esiobu, Fernandes, Fu, Fu, Fuller, Gao, Goswami, Goyal, Hartshorn, Hosseini, Hou, Inan, Kardas, Kerkez, Khabsa, Kloumann, Korenev, Koura, Lachaux, Lavril, Lee, Liskovich, Lu, Mao, Martinet, Mihaylov, Mishra, Molybog, Nie, Poulton, Reizenstein, Rungta, Saladi, Schelten, Silva, Smith, Subramanian, Tan, Tang, Taylor, Williams, Kuan, Xu, Yan, Zarov, Zhang, Fan, Kambadur, Narang, Rodriguez, Stojnic, Edunov, and Scialom}]{Touvron2023Llama2O}
Hugo Touvron, Louis Martin, Kevin~R. Stone, Peter Albert, Amjad Almahairi, Yasmine Babaei, Nikolay Bashlykov, Soumya Batra, Prajjwal Bhargava, Shruti Bhosale, Daniel~M. Bikel, Lukas Blecher, Cristian~Cant{\'o}n Ferrer, Moya Chen, Guillem Cucurull, David Esiobu, Jude Fernandes, Jeremy Fu, Wenyin Fu, Brian Fuller, Cynthia Gao, Vedanuj Goswami, Naman Goyal, Anthony~S. Hartshorn, Saghar Hosseini, Rui Hou, Hakan Inan, Marcin Kardas, Viktor Kerkez, Madian Khabsa, Isabel~M. Kloumann, A.~V. Korenev, Punit~Singh Koura, Marie-Anne Lachaux, Thibaut Lavril, Jenya Lee, Diana Liskovich, Yinghai Lu, Yuning Mao, Xavier Martinet, Todor Mihaylov, Pushkar Mishra, Igor Molybog, Yixin Nie, Andrew Poulton, Jeremy Reizenstein, Rashi Rungta, Kalyan Saladi, Alan Schelten, Ruan Silva, Eric~Michael Smith, R.~Subramanian, Xia Tan, Binh Tang, Ross Taylor, Adina Williams, Jian~Xiang Kuan, Puxin Xu, Zhengxu Yan, Iliyan Zarov, Yuchen Zhang, Angela Fan, Melanie Kambadur, Sharan Narang, Aurelien Rodriguez, Robert Stojnic, Sergey Edunov, and
  Thomas Scialom. 2023.
\newblock \href {https://api.semanticscholar.org/CorpusID:259950998} {Llama 2: Open foundation and fine-tuned chat models}.
\newblock \emph{ArXiv}, abs/2307.09288.

\bibitem[{Verma et~al.(2023)Verma, Jangra, Verma, and Saha}]{verma-etal-2023-large}
Yash Verma, Anubhav Jangra, Raghvendra Verma, and Sriparna Saha. 2023.
\newblock \href {https://doi.org/10.18653/v1/2023.eacl-main.263} {Large scale multi-lingual multi-modal summarization dataset}.
\newblock In \emph{Proceedings of the 17th Conference of the European Chapter of the Association for Computational Linguistics}, pages 3620--3632, Dubrovnik, Croatia. Association for Computational Linguistics.

\bibitem[{Wang et~al.(2024)Wang, Feng, Liu, and Wang}]{wang20242m}
Dongsheng Wang, Xiaoqin Feng, Zeming Liu, and Chuan Wang. 2024.
\newblock \href {https://doi.org/10.1007/s10489-024-05490-2} {{2M-NER}: Contrastive learning for multilingual and multimodal ner with language and modal fusion}.
\newblock \emph{Applied Intelligence}, 54(8):6252–6268.

\bibitem[{Wang et~al.(2022)Wang, Tian, Gui, Li, Wang, Yan, Chen, and Xiao}]{wang-etal-2022-wikidiverse}
Xuwu Wang, Junfeng Tian, Min Gui, Zhixu Li, Rui Wang, Ming Yan, Lihan Chen, and Yanghua Xiao. 2022.
\newblock \href {https://doi.org/10.18653/v1/2022.acl-long.328} {{W}iki{D}iverse: A multimodal entity linking dataset with diversified contextual topics and entity types}.
\newblock In \emph{Proceedings of the 60th Annual Meeting of the Association for Computational Linguistics (Volume 1: Long Papers)}, pages 4785--4797, Dublin, Ireland. Association for Computational Linguistics.

\bibitem[{Wu et~al.(2020)Wu, Petroni, Josifoski, Riedel, and Zettlemoyer}]{wu2019scalable}
Ledell Wu, Fabio Petroni, Martin Josifoski, Sebastian Riedel, and Luke Zettlemoyer. 2020.
\newblock \href {https://doi.org/10.18653/v1/2020.emnlp-main.519} {Scalable zero-shot entity linking with dense entity retrieval}.
\newblock In \emph{Proceedings of the 2020 Conference on Empirical Methods in Natural Language Processing (EMNLP)}, pages 6397--6407, Online. Association for Computational Linguistics.

\bibitem[{Xing et~al.(2023)Xing, Zhao, Wu, Li, Zhang, and Dai}]{Xing2023DRINDR}
Shangyu Xing, Fei Zhao, Zhen Wu, Chunhui Li, Jianbing Zhang, and Xinyu Dai. 2023.
\newblock \href {https://api.semanticscholar.org/CorpusID:263831560} {{DRIN}: Dynamic relation interactive network for multimodal entity linking}.
\newblock \emph{Proceedings of the 31st ACM International Conference on Multimedia}.

\bibitem[{Xu et~al.(2022)Xu, Chen, Shi, and Hu}]{xu2022enhancing}
Zhenran Xu, Yulin Chen, Senbao Shi, and Baotian Hu. 2022.
\newblock \href {https://doi.org/10.1007/978-3-031-17189-5_19} {Enhancing entity linking with contextualized entity embeddings}.
\newblock In \emph{Natural Language Processing and Chinese Computing: 11th CCF International Conference, NLPCC 2022, Guilin, China, September 24–25, 2022, Proceedings, Part II}, page 228–239, Berlin, Heidelberg. Springer-Verlag.

\bibitem[{Yao et~al.(2024)Yao, Wang, Chen, Wang, Liu, Xu, Yu, and Huang}]{yao2023ameli}
Barry Yao, Sijia Wang, Yu~Chen, Qifan Wang, Minqian Liu, Zhiyang Xu, Licheng Yu, and Lifu Huang. 2024.
\newblock \href {https://aclanthology.org/2024.eacl-long.172} {Ameli: Enhancing multimodal entity linking with fine-grained attributes}.
\newblock In \emph{Proceedings of the 18th Conference of the European Chapter of the Association for Computational Linguistics (Volume 1: Long Papers)}, pages 2816--2834, St. Julian{'}s, Malta. Association for Computational Linguistics.

\bibitem[{Zhang et~al.(2021)Zhang, Li, and Yang}]{zhang2021attention}
Li~Zhang, Zhixu Li, and Qiang Yang. 2021.
\newblock \href {https://doi.org/10.1007/978-3-030-73197-7_35} {Attention-based multimodal entity linking with high-quality images}.
\newblock In \emph{Database Systems for Advanced Applications: 26th International Conference, DASFAA 2021, Taipei, Taiwan, April 11–14, 2021, Proceedings, Part II}, page 533–548, Berlin, Heidelberg. Springer-Verlag.

\bibitem[{Zhong et~al.(2023)Zhong, Ding, Liu, Du, Jin, and Tao}]{zhong2023knowledge}
Qihuang Zhong, Liang Ding, Juhua Liu, Bo~Du, Hua Jin, and Dacheng Tao. 2023.
\newblock \href {https://doi.org/10.1109/TKDE.2023.3250499} {Knowledge graph augmented network towards multiview representation learning for aspect-based sentiment analysis}.
\newblock \emph{IEEE Trans. on Knowl. and Data Eng.}, 35(10):10098–10111.

\bibitem[{Zhou et~al.(2020)Zhou, Rijhwani, Wieting, Carbonell, and Neubig}]{zhou2020improving}
Shuyan Zhou, Shruti Rijhwani, John Wieting, Jaime Carbonell, and Graham Neubig. 2020.
\newblock \href {https://doi.org/10.1162/tacl_a_00303} {Improving candidate generation for low-resource cross-lingual entity linking}.
\newblock \emph{Transactions of the Association for Computational Linguistics}, 8:109--124.

\end{thebibliography}

\bibliographystyle{acl_natbib}

\appendix

\newpage
\section{Appendix}












    
\section*{Entity Linking Annotation Guidelines} \label{sec:entity_annotation_guidelines}

This section includes the complete set guidelines that were provided to annotators for the entity linking task

This annotation process/study is for a project as part of a research group at \textbf{Carnegie Mellon University}. We are looking for annotators to help us link entities/mentions in news articles to their corresponding entry in existing knowledge bases. You will be given a news article title and an associated image, which can be used in cases of ambiguity.

\subsubsection*{Terms and Concepts to Understand the Process Better}

\begin{itemize}
    \item \textbf{Mention/Entity:} 
        These are words of interest (e.g., names of persons, locations, and companies) that are mapped from an input text to corresponding unique entities in a target knowledge base. Identifying these entities in text helps computers interpret and map the mentions to corresponding real-world objects.

    \item \textbf{Knowledge Base:}
        It is a centralized database for storing information and data. It is a structured database of knowledge that can include facts, concepts, rules, and relationships about the entities in the world. We will be using \textbf{WikiData}, and you will see entities from there in the search bar while using the tool.

    \item \textbf{Entity Type:}
        We focus on the following types: \textit{Person, Location, Organization, Country, Events, Works, Miscellaneous}. Please refer to the \textit{Types of Entities} section for definitions and examples of these entities.

    \item \textbf{Reference Materials:}
        The full annotation guidelines and demonstration videos are present here: \href{https://docs.google.com/document/d/1kE_59JpLsieX-B7lzCOCDZdYy-7PsgqQGqawVQhPF_8/edit?tab=t.0}{guidelines}, \href{https://drive.google.com/drive/folders/16g-Pi4jIYHbkzBhRg9ZAjN7KzdFhJ7JV?usp=drive_link}{videos}. 
\end{itemize}

\subsubsection*{Types of Entities and Examples}

These are the types of entities we are looking to link. Refer to the examples and definitions in order to get a deeper understanding of an entity. You do not have to specifically mark the entity type in the tool, but we hope the below taxonomy helps you understand what all constitutes as an entity.

\hypersetup{
    urlcolor=blue,
    linkcolor=black
}

\subsection*{\textbf{Types of Entities}}

These are the types of entities (Table \ref{tab:entity-types}) we are looking to link. Refer to the examples and definitions in order to get a deeper understanding of an entity. You do not have to specifically mark the entity type in the tool, but we hope the below taxonomy helps you understand what all constitutes as an entity.

\begin{table*}[htbp!]
\centering
\caption{Entity Types, Descriptions, and Examples used in the annotation process.}
\label{tab:entity-types}
\rowcolors{2}{lightgray}{white}
\begin{tabular}{@{}p{0.15\linewidth} p{0.55\linewidth} p{0.25\linewidth}@{}}
\toprule
\textbf{Entity Type} & \textbf{Description} & \textbf{Examples} \\
\midrule
\textbf{PER: Person} &
Refers to a person, famous personality, leader etc. or a title &
\textit{Elon Musk, Michael Jackson, Chief Justice of Court} \\

\textbf{ORG: Organization} &
Any organized body of people that is formed for a goal/purpose. This includes companies, sports-teams. &
\textit{Apple (company), Real Madrid} \\

\textbf{COUNTRY: Country} &
Entity indicating a country &
\textit{Australia, China} \\

\textbf{LOC: Location} &
Includes all physical locations that can be located on a map (and are not a country). Examples: continents, cities, monuments, sites, universities, buildings. &
\textit{Eiffel Tower, River Nile, Supreme Court of India, Europe} \\

\textbf{EVENT: Events} &
Any entity that has duration or start/end time associated with it. &
\textit{World War 2, Mother’s Day} \\

\textbf{WORK: Works} &
Items associated with a specific profession, e.g., apps, products, movies, songs, etc. These could include work-specific jargons. &
\textit{iPhone-12, Titanic (the movie), Civic (car model)} \\

\textbf{MISC: Miscellaneous} &
Entities that are intuitively entities but do not fall into any above categories. Example: food items, animal species. &
\textit{Tacos, Nilgai, Titanic (the ship)} \\
\bottomrule
\end{tabular}
\end{table*}

\noindent
\\
\textbf{Refer to the following examples to understand what entities fall under what entity type:}

\begin{enumerate}[leftmargin=1.5em, itemsep=1em]

    \item \textit{``\hlWORK{iPhone 12} sales banned in \hlCOUNTRY{France} immediately, you may be surprised to hear the reason.''} \\
    \textcolor{gray}{Entities are \hlWORK{iPhone12} (WORK) and \hlCOUNTRY{France} (COUNTRY).}

    \item \textit{``\hlPER{Shohei Ohtani} joins \hlORG{Dodgers} for 10 years and \$700 million, the highest amount in \hlLOC{North American} professional sports history.''} \\
    \textcolor{gray}{Entities are \hlPER{Shohei Ohtani} (PER), \hlORG{Dodgers} (ORG), and \hlLOC{North American} (LOC).}

    \item \textit{``Why is a 'selfie' in \hlLOC{Berlin}, the capital of \hlCOUNTRY{Germany}, considered strange and selfish?''} \\
    \textcolor{gray}{Entities are \hlLOC{Berlin} (LOC) and \hlCOUNTRY{Germany} (COUNTRY).}

    \item \textit{``How big a success is the new malaria vaccine \hlMISC{R21}?''} \\
    \textcolor{gray}{\hlMISC{R21} is an entity which falls under MISC category.}

\end{enumerate}

\subsection*{\textbf{How to Access the website}}


We use the \textbf{Inception tool} for our annotation. You will be provided with the following three things:
\begin{enumerate}
    \item Website URL
    \item Assigned Documents
    \item Login Instructions
\end{enumerate}

\begin{figure}
    \centering
    \includegraphics[width=0.45\textwidth]{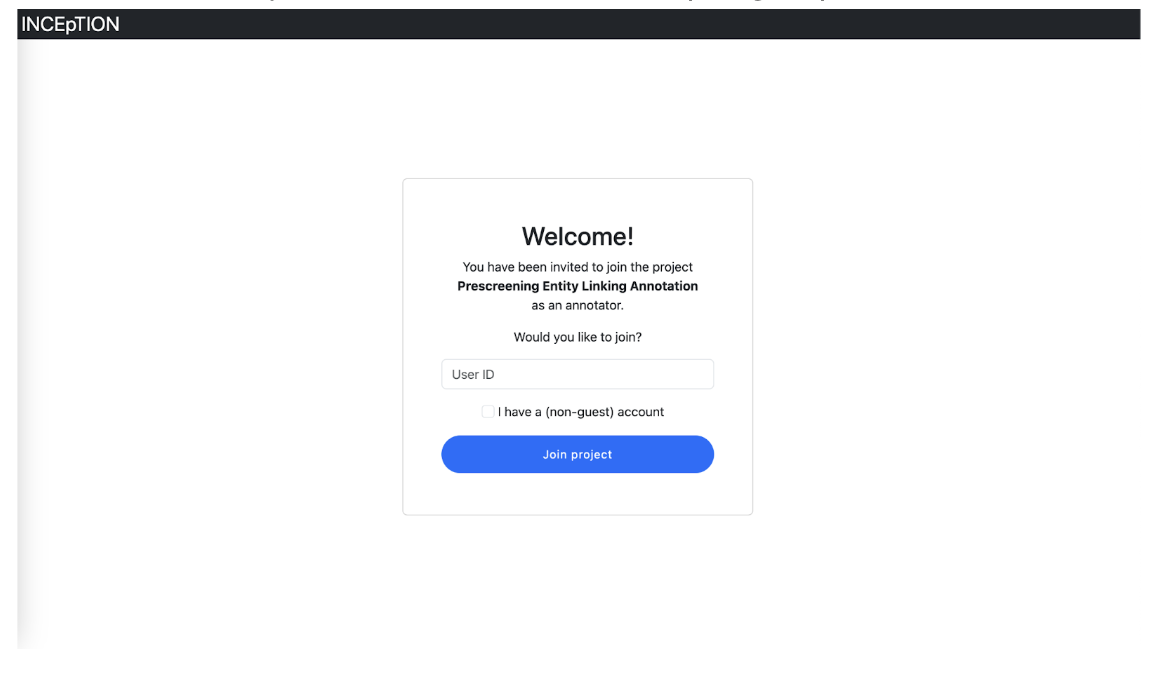}
    \caption{Login interface for the Inception annotation tool.}
    \label{fig:login}
\end{figure}

\begin{figure}
    \centering
    \includegraphics[width=0.45\textwidth]{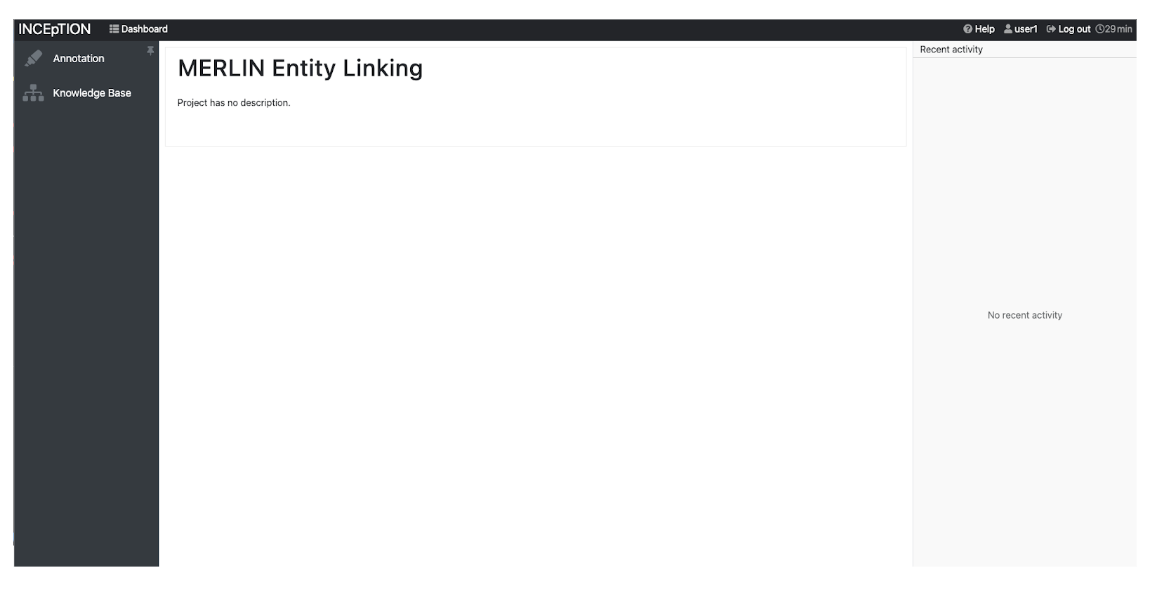}
    \caption{Inception tool home page interface.}
    \label{fig:homepage}
\end{figure}

Click on the URL in your preferred browser (Google Chrome is recommended). An interface like the one shown in \textbf{Figure \ref{fig:login}} will open. Enter a \textbf{User ID} to log in. Follow the naming convention suggested in the instructions. This will automatically create an account for you. \textbf{Do not} click on “I have a (non-guest) account”. This will take you to the home page as shown in \textbf{Figure \ref{fig:homepage}}.

\subsection*{Steps to Annotate}

\subsubsection*{1. Pick the First Document}

\begin{figure}
    \centering
    \includegraphics[width=0.45\textwidth]{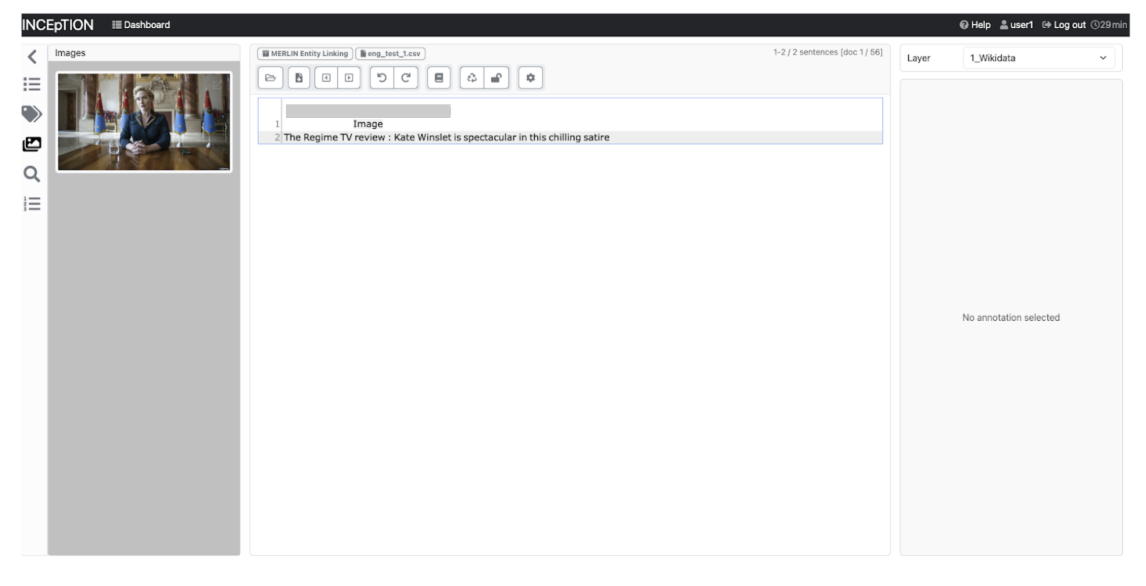}
    \caption{Document page view within the annotation tool.}
    \label{fig:document}
\end{figure}

Select the first document from the \textbf{Annotation tab}. Once selected, you will see the document page interface as in (Figure \ref{fig:document}).

\subsubsection*{2. Drag and Select Entities}

\begin{figure*}
    \centering
    \includegraphics[width=0.9\textwidth,height=6cm]{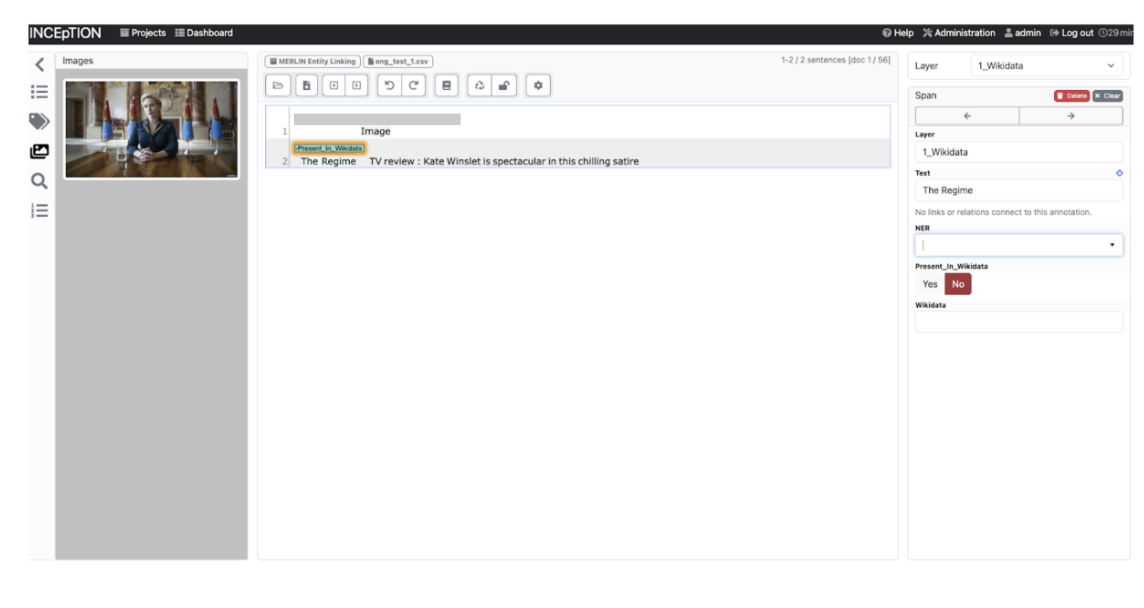}
    \caption{Selecting entity spans from text.}
    \label{fig:select}
\end{figure*}

There are two major components here, Image and Text (\textbf{Figure \ref{fig:select}}). Here it is an Image from the TV Show “The Regime” and it contains the text “The Regime TV review: Kate Winslet is spectacular in this chilling satire”

The text is interactive and you can drag and select portions of the text. This is the right way to select entities to annotate. For eg: You can drag and select mentions as shown in the image

\begin{figure*}[ht]
    \centering
    \includegraphics[width=0.9\textwidth,height=6cm]{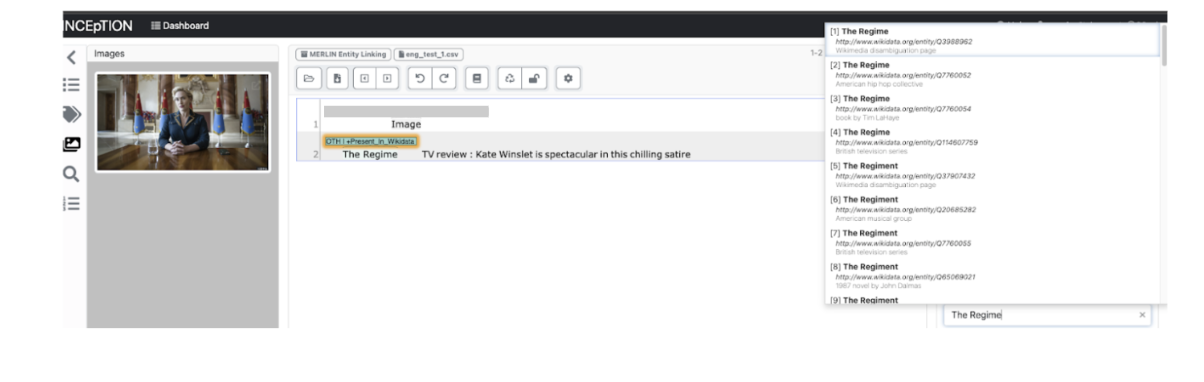}
    \caption{Searching, Disambiguating and Linking with the correct entity and type}
    \label{fig:regime}
\end{figure*}

\subsubsection*{3. Complete Annotation}

Now search for the entity in the \texttt{Wikidata} search bar as shown in \textbf{Figure \ref{fig:regime}} and pick the one which corresponds to the entity you think is appropriate for the text. \\

\noindent \textbf{Note}: "The Regime" is different from "Regime", so search accordingly. If needed, feel free to use google search with prompts like “The Regime Wikidata” or “Regime Wikidata” to identify the correct terms. Please refer to the \textbf{Effectively Search Wikidata \ref{sec:effective_search}} section for more details and cases. \\

\noindent Make sure to also pick the Entity Type in the \texttt{NER} text box. It is a dropdown menu and the final annotation will look like this. Please refer to the Types of Entities section to understand the differences across classes. \\

\noindent If you cannot find the entity in the search bar AND after doing a browser/google search, select “No” in \texttt{Present\_in\_Wikidata} and leave everything blank. This is a very rare case and in most cases you should be able to find the entity (if it is frequently used or famous). Make sure you change it to ‘Yes’ if you can find one. \\

\noindent Do this for ALL entity spans in the same document, eg: “Kate Winslet” is another entity in the same document, so repeat the same procedure. Once you have completed this document, navigate to the next document/annotation through the right arrow button at the top.

\subsection*{Effectively Searching Wikidata Entities}
\label{sec:effective_search}

In some cases, it might not be easy to find the correct entity directly from the wikidata search tab in the inception tool. Sometimes several entities with the same name would be returned which would need further digging to disambiguate and find the correct one. The search tab is also very sensitive to the textual query and it will only return the entity if there is an exact match in the name. We have listed some common cases below, please read through them thoroughly.

\begin{figure}[h]
    \centering
    \includegraphics[width=0.9\textwidth,height=6cm]{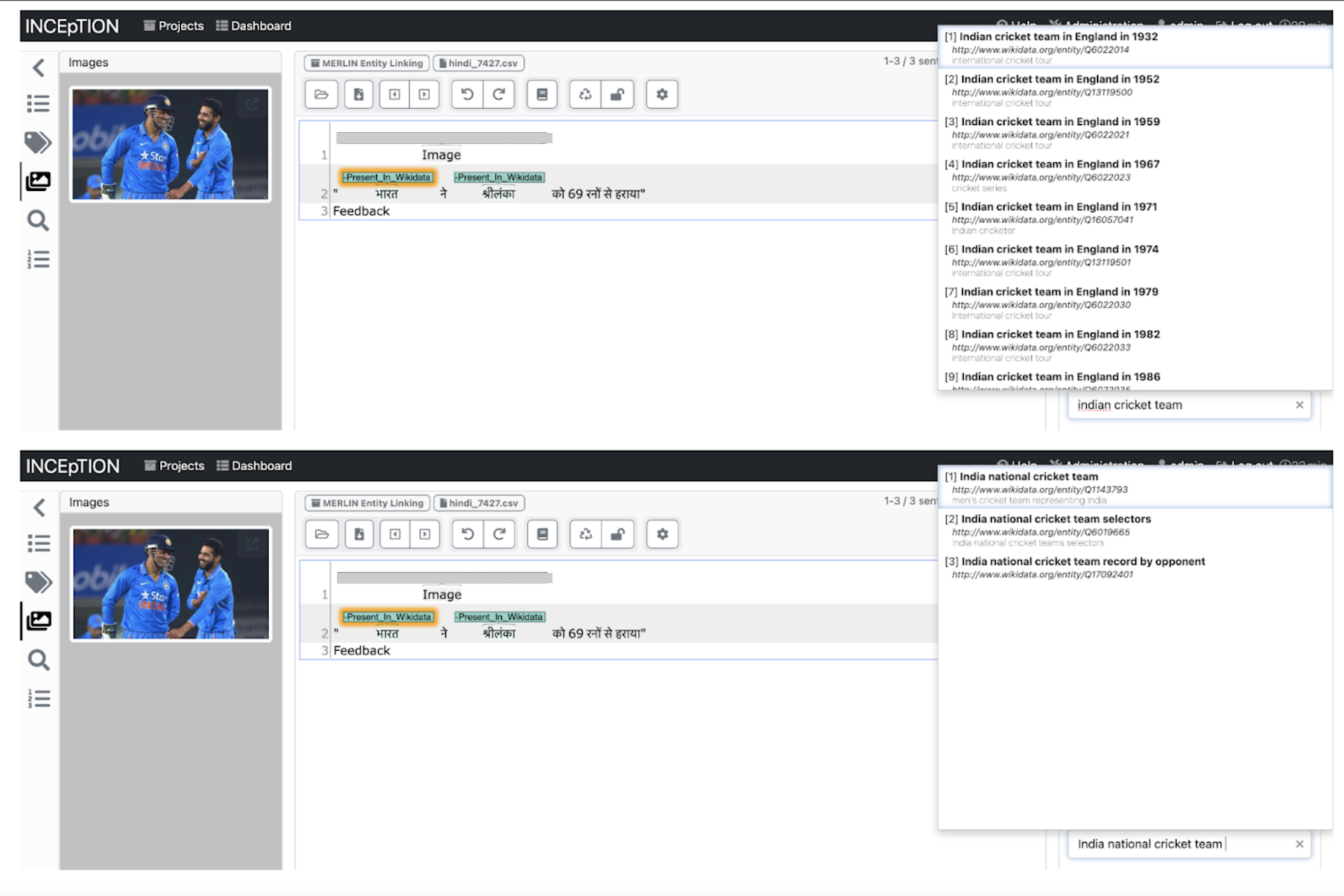}
    \caption{Example of disambiguating entities.}
    \label{fig:bharat}
\end{figure}

If you notice multiple entities in the search results, you will have to ensure that you select the correct entity. Here are some ways to identify those in cases of ambiguity: 

\begin{enumerate}
    \item \textbf{Search in English:} 

    Search in English: If searching from the wikidata tab on the left-hand side of your tool, your search query should be in English, so for instance if your entity is "\begin{CJK}{UTF8}{gbsn}  
埃菲尔铁塔
\end{CJK}" (English translation: Eiffel Tower), you will have to search for “Eiffel Tower".

    \item \textbf{Exact Word Sensitivity:}  
    Sometimes the entity’s Wikidata page name differs from your text.  
    Example: Transliteration: “Bharat ne Sri Lanka ko 69 runo se haraya” (Translation: “India beats Sri Lanka by 69 runs”), Here “Bharat / India” refers to the Indian cricket team.  
    Searching for “indian cricket team wikidata” in Google yields the correct page:  
    \url{https://www.wikidata.org/wiki/Q1143793}  
    The title there is “India national cricket team”, so that should be searched again in Inception (See \textbf{Figure \ref{fig:bharat}}).
\end{enumerate}

\begin{figure*}
    \centering
    \includegraphics[width=\textwidth]{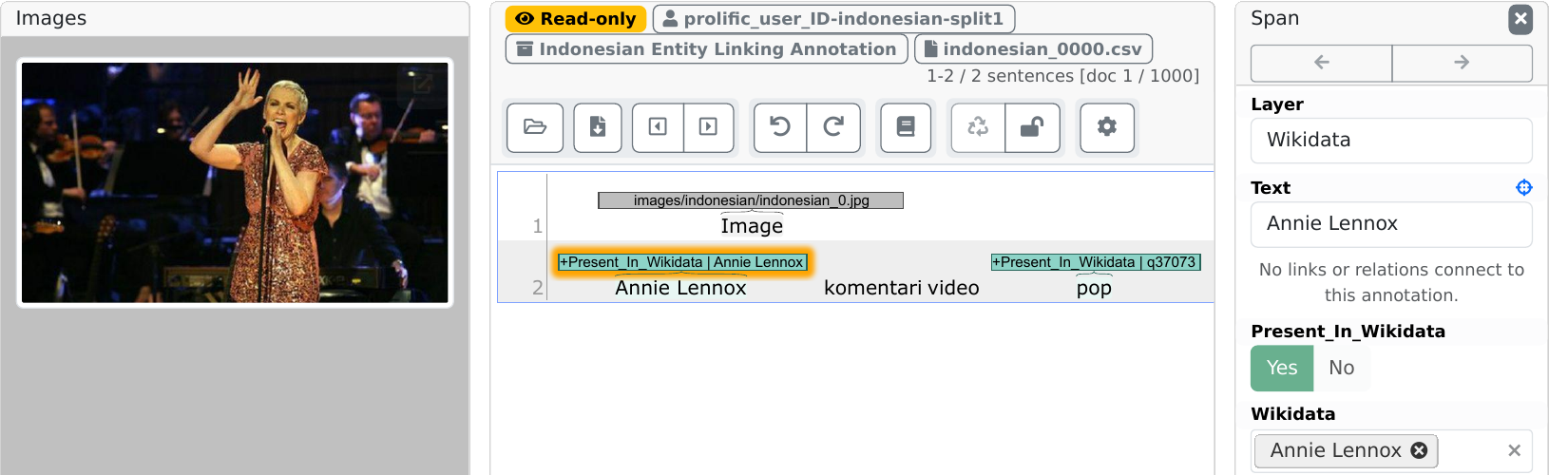}
    \caption{Entity Annotation in INCEpTION for Indonesian. The user selects and highlights text spans (e.g., "Annie Lennox") and assigns them to corresponding entities from a dropdown on the right side.}
    
    \label{fig:inception-example}
\end{figure*}

\begin{figure*}
    \centering
    \includegraphics[width=\linewidth]{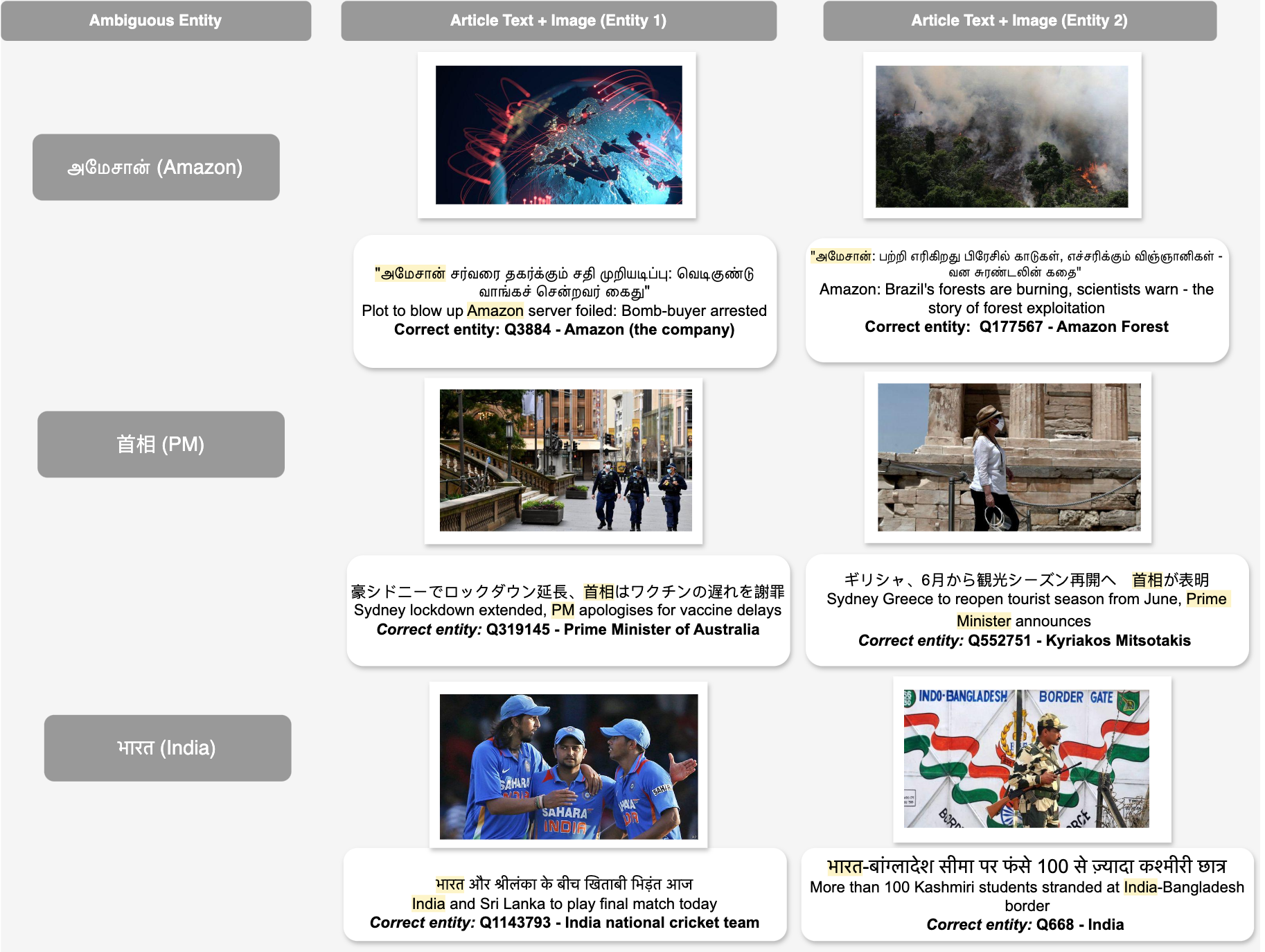} 
    \caption{\textit{Ambiguous hard mentions present in MERLIN} such as "Amazon," "PM," and "India" illustrate the challenge of entity disambiguation, where the same word can refer to multiple entities. For example, "Amazon" could refer to either the company or the rainforest, and "PM" might mean a Prime Minister from different countries. In these cases, combining images with textual context is crucial to accurately identify the correct entity. The visual context, such as the depiction of a forest or a tech server, narrows down the possible interpretations and we have such samples throughout the MERLIN dataset}
    \label{fig:ambig_entity_examples}

\end{figure*}

\begin{table*}
    \centering
    \begin{tabular}{l*{6}{c}} 
        \toprule
        \multirow{2}{*}{Model} & \multicolumn{6}{c}{Hindi (\%)} \\
        \cmidrule(lr){2-7}
        & PER & LOC & Country & ORG & MISC & Event \\
        \midrule
        GEMEL-Llama-2 & 58.69  & 47.48  & 84.04  & 45.95 \textuparrow & 16  & 36.51  \\
        GEMEL-Llama-2 w/o images & 62.22 & 57.89 & 92.18 & 43.24 & 28 & 45.24 \\
        GEMEL-Aya-23  & 72.80  & 57.28 \textuparrow  & 95.77 \textuparrow & 66.67  & 27.33 \textuparrow & 57.14  \\
        GEMEL-Aya-23  w/o images & 76.32 & 56.66 & 86.64 & 67.57 & 24.00 & 57.14 \\
        \midrule
        \multirow{2}{*}{Model} & \multicolumn{6}{c}{Indonesian (\%)} \\
        \cmidrule(lr){2-7}
        & PER & LOC & Country & ORG & MISC & Event \\
        \midrule
        GEMEL-Llama-2 & 87.54 \textuparrow & 62.81 \textuparrow & 83.86 \textuparrow & 80.54 \textuparrow & 44.44 \textuparrow & 51.77 \textuparrow \\
        GEMEL-Llama-2 w/o images & 74.02 & 57.19 & 77.78 & 68.46 & 34.44 & 46.81 \\
        GEMEL-Aya-23 & 82.92  & 63.16 \textuparrow & 87.42 \textuparrow & 73.15 \textuparrow & 50.00 \textuparrow & 60.28 \textuparrow \\
        GEMEL-Aya-23 w/o images & 83.63 & 56.84 & 79.66 & 72.48 & 48.89 & 50.35 \\
        \midrule
        \multirow{2}{*}{Model} & \multicolumn{6}{c}{Japanese (\%)} \\
        \cmidrule(lr){2-7}
        & PER & LOC & Country & ORG & MISC & Event \\
        \midrule
        GEMEL-Llama-2 & 63.31 \textuparrow & 75.59 \textuparrow & 90.69 \textuparrow & 70.65 \textuparrow & 55.77 \textuparrow & 53.55 \textuparrow \\
        GEMEL-Llama-2 w/o images & 54.44 & 63.73 & 86.20 & 68.46 & 34.44 & 46.81 \\
        GEMEL-Aya-23 & 73.67  & 76.95 \textuparrow & 69.66  & 69.15 \textuparrow & 48.08 \textuparrow & 62.09\textuparrow  \\
        GEMEL-Aya-23 w/o images & 75.15 & 72.54 & 78.33 & 67.66 & 44.23 & 38.66 \\
        \midrule
        \multirow{2}{*}{Model} & \multicolumn{6}{c}{Vietnamese (\%)} \\
        \cmidrule(lr){2-7}
        & PER & LOC & Country & ORG & MISC & Event \\
        \midrule
        GEMEL-Llama-2 & 71.01 \textuparrow & 75.88 \textuparrow & 75.57 \textuparrow & 65.15 \textuparrow & 39.73  & 43.16 \textuparrow \\
        GEMEL-Llama-2 w/o images & 64.71 & 64.98 & 72.31 & 60.61 & 43.84 & 38.95 \\
        GEMEL-Aya-23 & 67.23  & 65.37 \textuparrow & 76.71\textuparrow  & 65.15  & 49.32 \textuparrow & 57.89 \textuparrow \\
        GEMEL-Aya-23 w/o images & 72.27 & 64.59 & 73.13 & 68.18 & 38.36 & 44.21 \\
        \midrule
        \multirow{2}{*}{Model} & \multicolumn{6}{c}{Tamil (\%)} \\
        \cmidrule(lr){2-7}
        & PER & LOC & Country & ORG & MISC & Event \\
        \midrule
        GEMEL-Llama-2 & 24  & 22.11  & 30.79  & 13.19 \textuparrow & 3.33  & 6.61  \\
        GEMEL-Llama-2 w/o images & 29.78 & 28.57 & 45.53 & 10.42 & 12.22 & 15.70 \\
        GEMEL-Aya-23 & 30.67  & 37.76  & 53.42  & 33.33  & 7.78  & 31.40  \\
        GEMEL-Aya-23 w/o images & 42.67 & 42.52 & 66.05 & 49.31 & 16.67 & 35.54 \\
        \bottomrule
    \end{tabular}
    \caption{\textbf{R@1 Scores (in \%)for Models across Languages and Categories}: This table illustrates the varying influence of visual inputs across different mention types—PER (Person), LOC (Location), Country, ORG (Organization), MISC (Miscellaneous), and Event—across five languages: Hindi, Indonesian, Japanese, Vietnamese, and Tamil. Person mentions benefit most from visual context, particularly in disambiguating textually similar entities. Country and Organization mentions are generally easier for models to resolve, with minimal gains from images. However, MISC and Event mentions remain challenging, showing lower accuracy scores across models, highlighting the need for more sophisticated approaches to handle these types of mentions effectively. The up arrow (\textuparrow) indicates an improvement in accuracy when visual inputs are included for that method}
    \label{tab:category-performance}
\end{table*}

\end{document}